%% file: main.tex
\documentclass[10pt,twocolumn,letterpaper]{article}
\usepackage[pagenumbers]{cvpr}
\usepackage{algorithm}
\usepackage{algpseudocode}
\usepackage{float}
\usepackage{cuted}
\usepackage{xcolor}
\usepackage{booktabs}
\input{preamble}

\definecolor{linkblue}{rgb}{0.21,0.49,0.74}
\usepackage[breaklinks,colorlinks,allcolors=linkblue]{hyperref}
\hypersetup{
  pdftitle={TriDiT: Accelerating Video Editing via Dynamic Source Token Compression},
  pdfauthor={Han Wu and Songhua Liu},
  pdfsubject={Video editing and dynamic source token compression},
  pdfkeywords={video editing, diffusion transformers, token compression, efficient inference}
}
\title{TriDiT: Accelerating Video Editing via Dynamic Source Token Compression}
\author{%
Han Wu\textsuperscript{1}\qquad Songhua Liu\textsuperscript{2,*}\\[3pt]
{\normalsize \textsuperscript{1}Tianjin University}\\
{\normalsize \textsuperscript{2}Shanghai Jiao Tong University}\\[3pt]
{\small\texttt{wuhan2006@tju.edu.cn}\qquad\texttt{liusonghua@sjtu.edu.cn}}\\[2pt]
{\small \textsuperscript{*}Corresponding author}\\[3pt]
{\small \href{https://kimberley2006.github.io/TriDiT-page/}{Project page}\quad$\vert$\quad\href{https://github.com/Kimberley2006/TriDiT}{Code}}
}
\date{}
\makeatletter
\patchcmd{\@maketitle}{\vskip .375in}{\vskip 0pt}{}{}
\patchcmd{\@maketitle}{\vspace*{24pt}}{\vspace*{12pt}}{}{}
\patchcmd{\@maketitle}{\vskip .5em}{\vskip -.3em}{}{}
\makeatother

\begin{document}
\lefthyphenmin=2\righthyphenmin=3
\twocolumn[{%
  \renewcommand\twocolumn[1][]{#1}%
  \maketitle
  \input{sec/teaser_figure}
}]
\raggedbottom

\input{sec/0_abstract}
\input{sec/1_intro}
\input{sec/2_related}
\input{sec/3_method}
\input{sec/4_exp}
\input{sec/5_conclusion}
\section*{LLM Usage Statement}

We used ChatGPT to assist with manuscript drafting and revision,
the organization and clarification of methodological and
experimental descriptions, and LaTeX formatting.
LLM-generated suggestions do not constitute experimental evidence.
The authors remain responsible for the accuracy of all claims,
citations, and reported results, as well as the final content
of the manuscript.
{
    \small
    \bibliographystyle{ieeenat_fullname}
    \bibliography{main}
}
\input{sec/supplement}

\end{document}

%% file: preamble.tex
\usepackage{tikz}
\input{sec/qualitative_panels}

%% file: sec/qualitative_panels.tex
\definecolor{qualblue}{RGB}{0,111,190}
\definecolor{qualgreen}{RGB}{0,130,100}
\definecolor{qualorange}{RGB}{210,92,0}
\newcommand{\ROIImage}[3][]{%
\begin{tikzpicture}
\node[anchor=south west,inner sep=0] (im) at (0,0) {\includegraphics[#1]{#2}};
\begin{scope}[x={(im.south east)},y={(im.north west)}]
#3
\end{scope}
\end{tikzpicture}}
\newcommand{\Roi}[4]{%
\draw[#1,line width=0.7pt] #2 rectangle #3;
\node[anchor=north east,fill=white,fill opacity=.95,text opacity=1,
inner sep=1pt,font=\sffamily\bfseries\tiny,text=#1] at #3 {#4};}

%% file: sec/teaser_figure.tex
\begin{minipage}{\textwidth}
\centering
\includegraphics[width=0.96\textwidth]{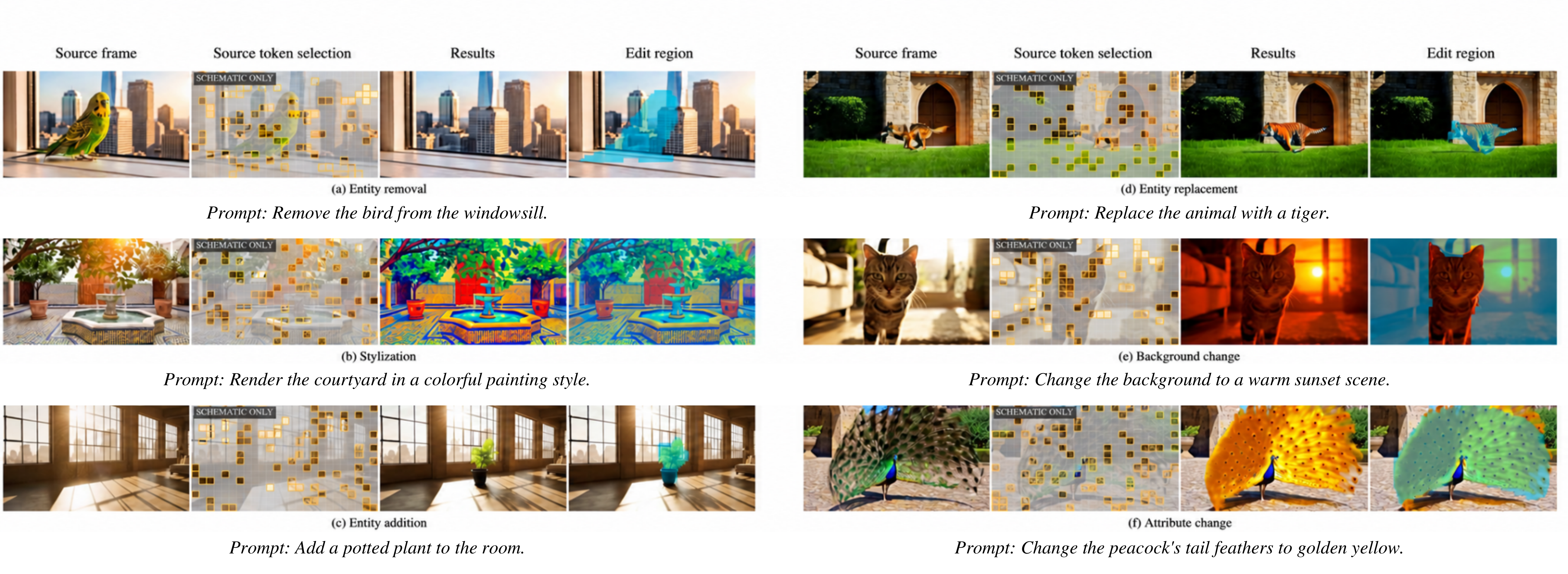}
\captionsetup{hypcap=false}
\captionof{figure}{\textbf{Illustrative editing categories and source-token selection.}
Illustrative prompts describe the displayed edits. Examples cover entity removal, stylization, entity addition, entity replacement,
background change, and attribute change.
Yellow cells schematically illustrate source-token selection; they are not recorded
representative IDs or measured attention maps. Cyan overlays indicate illustrated
edit regions. These static examples illustrate task types rather than temporal performance.}
\label{fig:teaser}
\end{minipage}
\vspace{10pt}

%% file: sec/0_abstract.tex
\begin{abstract}
Multimodal attention allows modern MMDiT-based models such as H3 to use a single architecture for video generation from text, images, or source videos. In source-conditioned video editing, however, repeatedly processing long source-token sequences across denoising steps and transformer layers incurs substantial computational overhead. To reduce this overhead, we propose TriDiT, a dynamic source-token compression framework for efficient video editing. Specifically, TriDiT formulates representative selection as a budget-constrained, demand-weighted facility-location problem that allocates representative capacity based on local content complexity, predicted attention importance, and irreplaceability. Because the resulting sparse coverage objective is monotone submodular, TriDiT can use exact lazy-greedy selection with the classical $1-1/e$ approximation guarantee. Once the representatives are selected, TriDiT aggregates the tokens assigned to each representative while preserving per-position residuals, with a default source-token retention ratio of $20\%$. The compressed source states are then updated across consecutive transformer layers. At each layer-group boundary, TriDiT uses the stored residuals to approximately reconstruct the source states at all positions before selecting a new set of representatives. By processing compressed source states within each layer group, TriDiT reduces source-side projection, attention, and feed-forward computation while retaining the full target layout. Experiments demonstrate that, on MiniMax-H3-Base-Ref2VA, TriDiT alone achieves a $2.49\times$ speedup in the DiT stage with a small reduction in editing quality. TriDiT is also compatible with existing sparse attention methods such as Spark Attention and regional feature reuse methods such as SpotEdit. When used together, the three methods achieve a $3.61\times$ DiT-stage speedup over the Dense baseline under the same seven-step inference configuration. Additional evaluations on UniVideo, UNIC, and Wan-VACE demonstrate that TriDiT applies across different source-conditioning interfaces.
\end{abstract}

%% file: sec/1_intro.tex
\section{Introduction}
Video-conditioned Diffusion Transformers (DiTs)~\cite{peebles2023dit}
support video generation and editing~\cite{bai2025recammaster,ma2024latte,yang2025cogvideox,wan2025wan,jiang2025vace},
using source appearance, structure, and temporal context to guide target
synthesis~\cite{cai2025ditctrl}. As duration and resolution increase,
repeatedly processing full source sequences incurs substantial attention,
projection, and feed-forward costs~\cite{dao2022flashattention,chen2026sparse,han2026veda}.
Yet source tokens contribute unequally: repeated appearance and geometry
create redundancy, while some local details can be approximated from
neighboring context.

Existing acceleration methods reduce repeated computation through
caching~\cite{zhao2025pab,lv2025fastercache,liu2025timestep}, sparse
attention such as Spark Attention~\cite{sparkattention2026}, or
region-aware execution and reuse, as in SpotEdit~\cite{qin2026spotedit}
and RegionE~\cite{chen2026regione}. Cross-frame feature propagation and source
conditioning also support coherent editing~\cite{geyer2024tokenflow,jiang2025vace}.
However, reducing interactions or reusing selected operations does not
necessarily shorten the source states processed throughout the network.
Token merging and selection offer a complementary route~\cite{bolya2022token,bolya2023token,li2024vidtome,lu2025toma}.
For source-conditioned editing, two issues motivate our design:
\emph{budget allocation} should reflect non-uniform content complexity,
attention demand, and substitutability; and \emph{state maintenance}
should retain positional distinctions so that previously merged positions
can participate in later selection.

We propose \textbf{TriDiT}, a dynamic source-compression framework for
video DiT editing. First, demand-weighted sparse coverage extends
coverage-based selection~\cite{lu2025toma} by combining predicted
attention, local reconstruction difficulty, and candidate substitutability.
Set-dependent marginal gains discourage redundant representatives~\cite{nemhauser1978analysis}.
Second, compact source sequences are updated across groups of five
Transformer layers. At each group boundary, token unmerging propagates
representative update increments to all original positions and combines
them with saved positional residuals before re-scoring and re-selection.
The target layout remains full, the generation backbone stays frozen,
and a lightweight offline-trained predictor estimates attention
demand~\cite{han2026veda}.

ToMA already combines facility-location selection, soft merging,
and cross-layer reuse~\cite{lu2025toma}; these ingredients alone are not
our novelty. Our focus is source-conditional demand weighting together
with continuous intra-group compression and boundary residual recovery.
This source-state reduction complements operator-level sparsity and
regional reuse. We therefore distinguish standalone compression gains,
controlled weighting gains, and benefits from external components.

Our contributions are:
\begin{enumerate}
    \item \textbf{Demand-aware budget allocation.} We integrate predicted
    attention, local complexity, and substitutability into sparse coverage
    demands, evaluated against uniform and single-factor controls.
    \item \textbf{Cross-layer compression with re-selection.} Short source
    sequences receive continuous updates; representative increments and
    positional residuals restore differentiated states for renewed budget
    competition at layer-group boundaries.
    \item \textbf{Quality--efficiency evaluation with separated attribution.}
    TriDiT achieves a reported $2.49\times$ standalone DiT speedup,
    $1.13\times$ over Attention-only at similar mean quality, and
    $3.61\times$ DiT-stage acceleration with Spark
    Attention~\cite{sparkattention2026} and SpotEdit~\cite{qin2026spotedit}.
\end{enumerate}

%% file: sec/2_related.tex
\section{Related Work}
\label{sec:related_work}

\noindent\textbf{Video Editing and Sparse Execution.}
Tune-A-Video uses one-shot tuning~\cite{wu2023tuneavideo}, FateZero fuses attention~\cite{qi2023fatezero}, and Rerender A Video propagates key-frame edits~\cite{yang2023rerender}.
TokenFlow propagates corresponding diffusion features~\cite{geyer2024tokenflow}; VACE unifies video generation and editing conditions~\cite{jiang2025vace}.
VSA and Veda learn sparse attention~\cite{zhang2025vsa,han2026veda}, Sliding Tile Attention uses local 3D windows~\cite{zhang2025sta}, and Sparse VideoGen exploits spatial--temporal head patterns~\cite{xi2025svg}.
LIVEditor combines context selection and query grouping~\cite{shao2026liveditor}.
SpotEdit selectively updates edited regions~\cite{qin2026spotedit}; RegionE combines adaptive partitioning, one-step unedited-region prediction, and caching~\cite{chen2026regione}.
TriDiT complements these approaches by shortening source states across layers.

\medskip
\noindent\textbf{Token Selection, Merging, and Reuse.}
Token Merging groups similar tokens~\cite{bolya2022token}; ToMeSD and VidToMe extend merging to diffusion and video editing~\cite{bolya2023token,li2024vidtome}.
ToFu combines pruning and merging with feature-norm correction~\cite{kim2024tofu}.
ToMA combines facility-location selection, soft merging, approximate recovery, and cross-step/layer reuse~\cite{lu2025toma}.
PAB, FasterCache, and TeaCache reuse denoising computations~\cite{zhao2025pab,lv2025fastercache,liu2025timestep}; AdaCache adapts caching schedules~\cite{kahatapitiya2025adacache}, and ToCa caches tokens selectively~\cite{zou2025toca}.
TriDiT instead continuously updates compact source states within layer groups, using tri-factor demand weights and boundary residual recovery to enable re-selection.

\medskip
\noindent\textbf{Comparison Scope.}
\textsc{ToMA-select} isolates demand weighting by using uniform weights with TriDiT's aggregation, recovery, and scheduling; it is not the full ToMA system~\cite{lu2025toma}.
Adaptation details are provided in Appendix~\ref{app:baselines}.

%% file: sec/3_method.tex
\section{Method}

\begin{figure*}[t]
  \centering
  \includegraphics[width=0.98\textwidth]{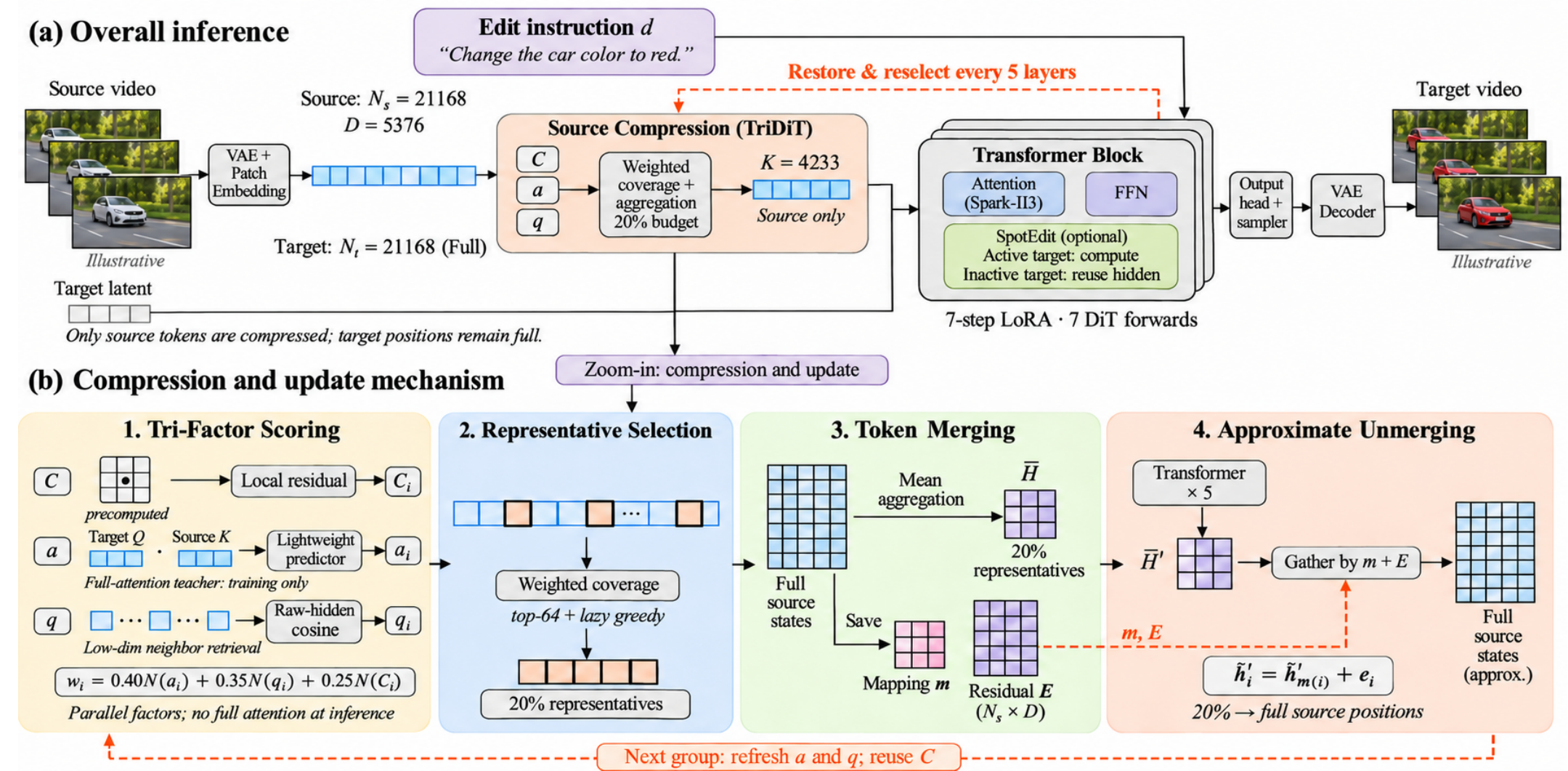}
  \caption{\textbf{Overview of TriDiT.}
  Tri-factor demands guide budget-constrained representative selection.
  Mean aggregation forms compact source states for consecutive layers;
  boundary residual recovery restores all source positions for re-selection.
  Target positions remain full throughout.}
  \label{fig:architecture}
\end{figure*}

\subsection{Problem Setup and Dynamic Source Compression}
\label{sec:source_pipeline}

Let $\mathcal S=\{1,\ldots,N_s\}$ index source tokens, with $N_t$
target positions and hidden dimension $D$.
At each selection boundary, TriDiT chooses representatives
$A\subseteq\mathcal S$ under the cardinality constraint $|A|\le K$,
where $K=\lfloor\rho N_s\rfloor$ and $\rho$ is the source retention
ratio.
This limits representative count rather than latency directly;
the target layout remains full.

By default, $\rho=0.20$ and each layer group contains
$L_{\mathrm{grp}}=5$ layers.
The backbone updates only the aggregated source representatives
within a group, reducing source-side projection, attention, and
FFN computation.
Selection starts from full source states at each denoising step;
subsequent groups use residual-based token unmerging to restore
approximate states for re-selection.
Figure~\ref{fig:architecture} summarizes the pipeline.
Tensor dimensions and operation counts are given in
Appendix~\ref{app:implementation}.

\subsection{Source-Token Retention Demands}
\label{sec:retention_demands}

Local complexity, predicted attention importance, and irreplaceability
measure reconstruction difficulty, target access demand, and substitute
scarcity, respectively.
They jointly define nonnegative coverage weights rather than
independent selection rankings.

\textbf{Local content complexity.}
Let $z_i$ be a fixed source latent patch descriptor,
$\bar z_i^{(s)}$ the mean of its valid four-neighbors at the same
temporal slice with spacing $s\in\{1,2\}$, and $d_z$ its dimension.
We define
\begin{equation}
C_i=\frac12\sum_{s\in\{1,2\}}
\frac{\|z_i-\bar z_i^{(s)}\|_2^2/d_z}
{\|z_i\|_2^2/d_z+\|\bar z_i^{(s)}\|_2^2/d_z+\epsilon}.
\label{eq:complexity}
\end{equation}
Computed once per generation, $C_i$ measures local reconstruction
difficulty rather than motion or semantic importance directly.

\textbf{Irreplaceability.}
At a selection boundary, let $\tilde h_i$ denote the full-position
source representation: the initialized state at the first group,
or an approximately restored state thereafter.
Low-dimensional retrieval constructs a nonempty spatiotemporal
candidate set $\mathcal C(i)$ excluding $i$.
We compute
\begin{equation}
\begin{aligned}
s_{ij}
&=\frac{\tilde h_i^\top\tilde h_j}
{\max(\|\tilde h_i\|_2,\epsilon)
 \max(\|\tilde h_j\|_2,\epsilon)},\\
q_i
&=\frac12\left(1-\max_{j\in\mathcal C(i)}s_{ij}\right).
\end{aligned}
\label{eq:substitutability}
\end{equation}
A larger $q_i$ indicates fewer close substitutes in the candidate set.
Similarities are refreshed at selection boundaries and remain fixed
during each selection round.

\textbf{Predicted attention importance.}
Following attention-prediction ideas in sparse diffusion
acceleration~\cite{han2026veda}, we partition full target queries
into $G=8$ consecutive blocks.
For head $h$, let $T_g^{Q,h}$ concatenate the elementwise average,
maximum, and minimum query statistics in block $B_g$, with
$\omega_g=|B_g|/N_t$.
Head-specific projections predict source attention demand:
\begin{equation}
\begin{aligned}
\hat q_g^h
&=W_q^hT_g^{Q,h}+b_q^h,\qquad
\hat k_i^h=W_k^h k_i^{S,h}+b_k^h,\\
\widehat P_{gi}^h
&=\operatorname{softmax}_{i\in\mathcal S}
\left((\hat q_g^h)^\top\hat k_i^h/\sqrt{d'}\right),\\
a_i
&=H_a^{-1}\sum_{h,g}\omega_g\widehat P_{gi}^h.
\end{aligned}
\label{eq:predicted_attention}
\end{equation}
The predictor shares parameters across layers and denoising steps
and estimates demand for the next layer.
It is trained offline on uncompressed trajectories with the backbone,
LoRA~\cite{hu2022lora}, and teacher frozen; inference avoids full
teacher-attention evaluation.
Later groups use restored states, introducing a train--inference
discrepancy.
Training details and cache handling are given in
Appendix~\ref{app:implementation}; all predictor and maintenance
costs are included in latency.

\textbf{Demand weights.}
Let $\mathcal N(x)_i=x_i/\sum_{j\in\mathcal S}x_j$ for a nonzero
nonnegative score vector, with $\mathcal N(x)=0$ for an all-zero
vector.
We define
\begin{equation}
\begin{aligned}
w_i
&=\lambda_a\mathcal N(a)_i
 +\lambda_q\mathcal N(q)_i
 +\lambda_c\mathcal N(C)_i,\\
(\lambda_a,\lambda_q,\lambda_c)
&=(0.40,0.35,0.25).
\end{aligned}
\label{eq:weights}
\end{equation}
Only $a_i$ explicitly depends on the current target.
Irreplaceability $q_i$ weights demand, whereas coverage similarity
measures how well a representative serves it.
The factors need not be statistically independent; their incremental
utility is assessed in Sec.~4.3.

\emph{Evidence for local proxies.}
Local probes support associations with reconstruction, deletion,
and replacement sensitivity, rather than full-generation quality;
protocols, results, and limitations are reported in
Appendix~\ref{app:factors}.

\subsection{Budgeted Demand-Weighted Coverage Maximization}
\label{sec:coverage_selection}

\textbf{Facility-location formulation.}
Each source token is both a demand point and a candidate representative.
Scoring each position by its best selected representative yields a
demand-weighted facility-location objective under a count
budget~\cite{lu2025toma}.

Low-dimensional retrieval reuses predictor projections.
We partition space into $6\times6$ blocks and search the containing
block, adjacent blocks including diagonals, and neighboring temporal
slices, then rerank candidates using original hidden-state cosine
similarity.
Each representative candidate $u$ has a sparse neighborhood
$\mathcal E(u)$ with at most 64 neighbors and self-coverage.
Define
\begin{equation}
R_{ui}=\max(0,s_{ui}),\quad R_{uu}=1,\quad
\widetilde R_{ui}=R_{ui}\mathbf 1[i\in\mathcal E(u)].
\label{eq:coverage_graph}
\end{equation}
The representative selection problem is
\begin{equation}
\begin{aligned}
A^\star
&\in\arg\max_{A\subseteq\mathcal S,\,|A|\le K}
\widetilde F(A),\\
\widetilde F(A)
&=\sum_{i\in\mathcal S}w_i
\max_{u\in A}\widetilde R_{ui},
\qquad \widetilde F(\varnothing)=0.
\end{aligned}
\label{eq:coverage}
\end{equation}
Weights and graph edges remain fixed throughout one selection round.
The graph may be directed; original source indices identify all
positions.
Retrieval takes $2$--$3\,\mathrm{ms}$ in the reported configuration
and is included in maintenance overhead.

\textbf{Why weighted coverage?}
Let $b_i(A)=\max_{u\in A}\widetilde R_{ui}$, with
$b_i(\varnothing)=0$.
The weight $w_i$ encodes demand, while $b_i(A)$ measures how well
the current set serves that demand.
Once a position is well covered, a similar additional representative
contributes little extra value.
Equivalently, maximizing Eq.~\eqref{eq:coverage} minimizes the
weighted under-coverage proxy $\sum_i w_i[1-b_i(A)]$ under the
same count budget.
This is not a minimum-cardinality full-cover problem, nor an upper
bound on generation error.

\textbf{Maximum-marginal-gain selection.}
Starting from $A=\varnothing$, each iteration selects the source
token that produces the largest increase in the weighted coverage
objective:
\begin{equation}
\begin{aligned}
\Delta(u\mid A)
&=\widetilde F(A\cup\{u\})-\widetilde F(A)\\
&=\sum_{i\in\mathcal E(u)}w_i
\max\bigl(0,R_{ui}-b_i(A)\bigr),\\
u^\star
&\in\arg\max_{u\in\mathcal S\setminus A}\Delta(u\mid A),
\qquad A\leftarrow A\cup\{u^\star\}.
\end{aligned}
\label{eq:marginal_gain}
\end{equation}
After each addition, coverage is updated as
$b_i\leftarrow\max(b_i,\widetilde R_{u^\star i})$.
Selection continues until $|A|=K$, breaking ties by original
source index.
Thus the selection score is the marginal gain $\Delta(u\mid A)$,
not the fixed demand weight $w_u$.

Exact lazy greedy~\cite{minoux1978accelerated} caches marginal-gain
upper bounds and recomputes the leading candidate, accepting it only
when its exact gain dominates all remaining bounds.
Otherwise, its bound is updated and the heap revisited, preserving
valid greedy choices without heuristic truncation.

\textbf{Proposition 1 (Approximation guarantee).}
\emph{For fixed nonnegative weights and fixed nonnegative sparse
coverage relations, $\widetilde F$ is normalized, monotone, and
submodular.
Under the sole constraint $|A|\le K$, with $1\le K\le N_s$,
standard greedy and its exact lazy implementation return a set
$A_{\mathrm g}$ satisfying}
\begin{equation}
\begin{aligned}
\widetilde F(A_{\mathrm g})
&\ge
\left[1-\left(1-\frac1K\right)^K\right]
\widetilde F(A^\star)\\
&\ge(1-1/e)\widetilde F(A^\star).
\end{aligned}
\label{eq:approximation_guarantee}
\end{equation}

\emph{Proof.}
Normalization follows by definition, and
Eq.~\eqref{eq:marginal_gain} is nonnegative.
For $A\subseteq B$ and $u\notin B$, $b_i(A)\le b_i(B)$ implies
$\Delta(u\mid A)\ge\Delta(u\mid B)$, establishing submodularity.
The classical cardinality-constrained greedy bound then gives
Eq.~\eqref{eq:approximation_guarantee}~\cite{nemhauser1978analysis}.
Submodularity also makes cached earlier marginal gains valid upper
bounds, so exact lazy evaluation preserves the guarantee.

This guarantee concerns the optimum of the same sparse objective
within one selection round; it does not extend directly to a dense
coverage graph, generation quality, or the complete sequence of
re-selection decisions.
Refreshing weights and graph edges between layer groups defines
a new optimization instance.
Setting $w_i$ to a uniform positive constant gives our controlled
\textsc{ToMA-select} baseline, with the other pipeline components
unchanged; it is not the full original ToMA
implementation~\cite{lu2025toma}.

\subsection{Mean Aggregation and Residual-Based Token Unmerging}
\label{sec:token_unmerging}

\textbf{Assignment and aggregation.}
After selecting $K$ representatives, each representative maps to
itself.
Every other source token maps to the selected representative with
the highest original cosine similarity among its incoming coverage
edges.
If none exists, we compare it with all selected representatives.
Let $m(i)\in A$ denote this assignment and
$\mathcal G_u=\{i:m(i)=u\}$ its token group.
These token groups are distinct from the groups of Transformer layers.

At the default 20\% retention ratio, the remaining 80\% of source
tokens are assigned to representatives, up to integer rounding.
Each token group, including its representative, is aggregated by
an unweighted mean~\cite{li2024vidtome,liu2025timestep},
with per-position residuals saved separately:
\begin{equation}
\begin{aligned}
\bar h_u
&=\frac1{|\mathcal G_u|}
\sum_{i\in\mathcal G_u}\tilde h_i,\\
e_i
&=\tilde h_i-\bar h_{m(i)},\qquad
\tilde h_i'=\bar h_{m(i)}'+e_i.
\end{aligned}
\label{eq:recovery}
\end{equation}
Here $\bar h_u'$ is the representative state after the layer group,
and the last equality defines token unmerging.
The demand weights $w_i$ affect selection, not aggregation.
Representatives retain their original spatiotemporal coordinates.
Fallback assignment does not change the selection objective.

\textbf{Compressed execution and unmerging.}
The backbone updates the $K$ aggregated source representatives
within each layer group, while assignments and residuals remain
fixed.
Before the next group, token unmerging restores approximate states
at all $N_s$ source positions using Eq.~\eqref{eq:recovery}.
Equivalently,
\begin{equation}
\tilde h_i'=\tilde h_i+\Delta\bar h_{m(i)},\qquad
\Delta\bar h_u=\bar h_u'-\bar h_u.
\label{eq:increment}
\end{equation}
Each source position receives its representative's update increment.
For two positions assigned to the same representative,
$\tilde h_i'-\tilde h_j'=e_i-e_j$, preserving their pre-aggregation
difference.
Residuals do not evolve independently inside the group, so this
operation does not reconstruct the exact dense trajectories.
Saving full-position residuals also means that 20\% representative
retention does not imply an 80\% reduction in total stored source
information.

\textbf{Periodic re-selection.}
Following unmerging, we refresh $a$ and $q$, reuse $C$, rebuild the
sparse graph, and select $K$ representatives by maximum marginal gain.
New assignments and residuals replace the previous records.
All restored positions are eligible, including those merged in
the preceding group.
Re-selection changes which positions share future updates; it does
not guarantee monotonically decreasing generation error.
The final layer group is unmerged only if required by a downstream
interface (Appendices~\ref{app:implementation} and~\ref{app:recovery}).

Algorithm~\ref{alg:tridit} summarizes inference for the target-output
interface.
Here $\mathsf G$ denotes the sparse coverage graph, $R$ the
representative-state matrix, and $E$ the residual matrix.
$\mathrm{RunGroup}$ executes both the compressed source sequence
and the target branch; source compression does not remove target
positions.

\begin{algorithm}[!t]
\caption{TriDiT Inference}
\label{alg:tridit}
\begin{algorithmic}[1]
\Require Source $Z_s$, target noise $Z_T$, condition $d$;
         retention $\rho$, layer-group size $L_{\mathrm{grp}}$
\Ensure Edited target latent $Z_0$
\State $C\gets\mathrm{Complexity}(Z_s)$;
       $K\gets\lfloor\rho N_s\rfloor$
\For{$t=T,T-1,\ldots,1$}
  \State $(H_s,H_t,c)\gets
         \mathrm{Initialize}(Z_s,Z_t,d,\tau_t)$
  \For{$b=0,\ldots,\lceil L/L_{\mathrm{grp}}\rceil-1$}
    \State $H_s\gets R[m]+E$ \textbf{if} $b>0$
           \Comment{Token unmerging}
    \State $(a,q,\mathsf G)\gets
           \mathrm{ScoreAndGraph}(H_s,H_t,c)$
    \State $w\gets\lambda_a\mathcal N(a)
                   +\lambda_q\mathcal N(q)
                   +\lambda_c\mathcal N(C)$
    \State $A\gets\mathrm{LazyGreedyCoverage}(\mathsf G,w,K)$
    \State $(R,m,E)\gets\mathrm{MeanMerge}(H_s,A,\mathsf G)$
    \State $(R,H_t)\gets\mathrm{RunGroup}_b(R,H_t,c;A)$
  \EndFor
  \State $Z_{t-1}\gets\mathrm{OutputAndSample}(Z_t,H_t,c)$
\EndFor
\State \Return $Z_0$
\end{algorithmic}
\end{algorithm}

\subsection{Component Interfaces and Scope of Applicability}
\label{sec:component_interfaces}

Spark Attention (implemented by Spark-H3) sparsifies attention interactions over the representative-compressed sequence.
With SpotEdit~\cite{qin2026spotedit}, active target positions are
updated and inactive positions reuse cached states, retaining
the full target layout and logical K/V context.
Original coordinates and visibility rules are preserved.

Full source compression requires independently indexed source
positions and operators supporting shortened sequences.
UniVideo~\cite{wei2026univideo} and UNIC~\cite{ye2026unic} use
joint-sequence interfaces with backbone-specific predictor
training or calibration.
Wan-VACE~\cite{jiang2025vace} is evaluated separately because its
conditional branch changes the compression scope.
Interface definitions, cache rules, and restoration requirements
are detailed in Appendix~\ref{app:implementation}.

%% file: sec/4_exp.tex
%

\setcounter{dbltopnumber}{1}

\begin{table*}[!t]
\centering
\scriptsize
\setlength{\tabcolsep}{2.2pt}
\renewcommand{\arraystretch}{1.07}
\begin{tabular*}{\textwidth}{@{\extracolsep{\fill}}lrrrrrrrrrr@{}}
\toprule
& Preservation & \multicolumn{4}{c}{GT reference} & \multicolumn{3}{c}{Dense reference} & \multicolumn{2}{c}{Efficiency} \\
\cmidrule(lr){2-2}\cmidrule(lr){3-6}\cmidrule(lr){7-9}\cmidrule(l){10-11}
Method & SSIM-S$\uparrow$ & PSNR$\uparrow$ & SSIM$\uparrow$ & LPIPS$\downarrow$ & FID$\downarrow$ & PSNR$\uparrow$ & SSIM$\uparrow$ & LPIPS$\downarrow$ & DiT/s$\downarrow$ & Speedup$\uparrow$ \\
\midrule
\multicolumn{11}{@{}l}{\textbf{External video editing systems}} \\
InsV2V & 0.901 & 25.12 & 0.842 & 0.208 & 52.6 & 25.10 & 0.840 & 0.210 & 41.70 & $0.73\times$ \\
LGVI & 0.889 & 24.58 & 0.831 & 0.226 & 58.3 & 24.50 & 0.830 & 0.228 & 48.30 & $0.63\times$ \\
Lucy-Edit-dev & 0.925 & 27.31 & 0.876 & 0.155 & 40.8 & 27.35 & 0.875 & 0.158 & 24.50 & $1.24\times$ \\
VACE & 0.918 & 26.54 & 0.864 & 0.173 & 44.7 & 26.50 & 0.860 & 0.178 & 32.80 & $0.93\times$ \\
Ditto & 0.931 & 27.65 & 0.881 & 0.149 & 39.5 & 27.70 & 0.880 & 0.152 & 23.40 & $1.30\times$ \\
FastH3-Omni & 0.922 & 27.08 & 0.872 & 0.161 & 42.1 & 27.10 & 0.870 & 0.165 & 12.79 & $2.38\times$ \\
\midrule
\multicolumn{11}{@{}l}{\textbf{Same-backbone baselines}} \\
Dense & 0.936 & 27.84 & 0.884 & 0.146 & 39.2 & -- & -- & -- & 30.45 & $1.00\times$ \\
Spark-H3 & 0.931 & 27.59 & 0.880 & 0.151 & 40.1 & 35.42 & 0.976 & 0.028 & 17.10 & $1.78\times$ \\
SpotEdit & 0.940 & 27.88 & 0.885 & 0.145 & 39.4 & 34.76 & 0.972 & 0.032 & 25.60 & $1.19\times$ \\
Spark-H3 + SpotEdit & 0.934 & 27.58 & 0.880 & 0.152 & 40.5 & 33.87 & 0.965 & 0.041 & 14.80 & $2.06\times$ \\
\midrule
\multicolumn{11}{@{}l}{\textbf{Standalone source compression, 20\% retention}} \\
ToMe adaptation & 0.910 & 26.43 & 0.861 & 0.180 & 46.8 & 29.24 & 0.927 & 0.083 & 11.85 & $2.57\times$ \\
VidToMe adaptation & 0.916 & 26.71 & 0.867 & 0.172 & 44.9 & 30.18 & 0.938 & 0.071 & 12.05 & $2.53\times$ \\
Full ToMA adaptation & 0.925 & 27.22 & 0.875 & 0.160 & 42.3 & 31.86 & 0.953 & 0.054 & 12.60 & $2.42\times$ \\
ToMA-select & 0.915 & 26.62 & 0.865 & 0.176 & 45.6 & 29.87 & 0.934 & 0.076 & 12.10 & $2.52\times$ \\
\textbf{TriDiT} & \textbf{0.932} & \textbf{27.63} & \textbf{0.881} & \textbf{0.151} & \textbf{40.2} & \textbf{33.64} & \textbf{0.967} & \textbf{0.039} & 12.229 & $2.49\times$ \\
\midrule
\multicolumn{11}{@{}l}{\textbf{Combined acceleration}} \\
\textbf{TriDiT + SpotEdit} & 0.937 & 27.60 & 0.880 & 0.152 & 40.6 & 32.98 & 0.962 & 0.045 & 10.20 & $2.99\times$ \\
\textbf{TriDiT + Spark-H3 + SpotEdit} & 0.930 & 27.41 & 0.877 & 0.157 & 41.5 & 31.92 & 0.952 & 0.055 & \textbf{8.435} & $\mathbf{3.61}\times$ \\
\bottomrule
\end{tabular*}
\caption{\textbf{Quality, preservation, Dense fidelity, and efficiency.}
SSIM-S evaluates non-edited source regions; GT and Dense columns use edited
references and matching Dense outputs (PSNR in dB).
Same-backbone timing uses four A100 80GB GPUs, seven-step LoRA, CFG disabled,
and 50 tasks. DiT latency includes maintenance/component overheads and excludes
text encoding/VAE processing; speedup is $30.45\,\mathrm{s}/T_{\mathrm{method}}$.
``--'' denotes unreported/inapplicable comparisons, including Dense self-comparison.
Bold quality scores are best among standalone compression methods;
bold efficiency scores are best overall. Protocols: Appendices~\ref{app:metrics} and~\ref{app:timing}.}
\label{tab:main_results}
\end{table*}

\section{Experiments}
\label{sec:experiments}

\subsection{Experimental Setup and Protocols}
\label{sec:experimental_setup}

Our experiments use MiniMax-H3-Base-Ref2VA with four
NVIDIA A100 80GB GPUs jointly performing single-video
inference. All same-backbone comparisons use a seven-step
distillation LoRA with CFG disabled.
Unless otherwise stated, source retention is $20\%$
and re-selection occurs every $L_{\mathrm{grp}}=5$ layers.
Additional evaluations on UniVideo~\cite{wei2026univideo},
UNIC~\cite{ye2026unic}, and Wan-VACE~\cite{wan2025wan,jiang2025vace}
assess applicability across source-conditioning interfaces;
transfer configurations are detailed in Appendix~\ref{app:evaluation}.

The test set contains 100 videos and 400 tasks.
Three raters assess editing completion and temporal
consistency. Table~\ref{tab:main_results} separates source-preservation
SSIM-S, GT-referenced quality, fidelity to Dense, and inference efficiency.
The masked Dense evaluation covers 380 non-style tasks;
this subset does not define the scope of every metric in Table~\ref{tab:main_results}.
Dataset provenance, task construction, data splits,
scoring protocols, and evaluation reliability are
detailed in Appendix~\ref{app:evaluation}.

Efficiency is evaluated on a fixed subset of 50 tasks.
DiT latency includes compression maintenance and
component overheads, but excludes text encoding and
VAE processing.
Timing and memory protocols are detailed in
Appendix~\ref{app:timing}; baseline adaptations
and their comparison scope are specified in
Appendix~\ref{app:baselines}.

\subsection{Benefits and Attribution of the Compression Pipeline}
\label{sec:pipeline_benefits}

In Table~\ref{tab:main_results}, standalone TriDiT improves over
Full ToMA adaptation at the same 20\% retention: SSIM-S rises from
0.925 to 0.932, GT PSNR from 27.22 to 27.63\,dB, and Dense LPIPS
falls from 0.054 to 0.039. Relative to Dense, GT PSNR decreases by
0.21\,dB and GT LPIPS increases by 0.005. The full combination has
lower GT PSNR (27.41\,dB) and higher Dense LPIPS (0.055) than
standalone TriDiT, exposing a quality--efficiency trade-off.

On the 50-task efficiency subset, standalone TriDiT reduces DiT
latency from $30.45$\,s to $12.229$\,s ($2.49\times$ speedup;
Table~\ref{tab:main_results}), with editing and temporal scores
decreasing by 1.0 and 1.4 points.
This comparison evaluates the complete compression
pipeline; the contribution of demand weighting is
isolated in Sec.~\ref{sec:demand_weighting}.

With Spark-H3 and SpotEdit~\cite{qin2026spotedit},
TriDiT reaches a reported $3.61\times$ DiT speedup.
Relative to Spark-H3 + SpotEdit, latency decreases
from $14.80$\,s to $8.435$\,s, an incremental
$1.75\times$ speedup, while editing completion
increases by 1.5 points.
These are denoising-stage rather than end-to-end gains.
Additional comparisons are provided in
Appendix~\ref{app:baselines}.

\begin{figure}[!t]
    \centering
    \includegraphics[width=\columnwidth]
        {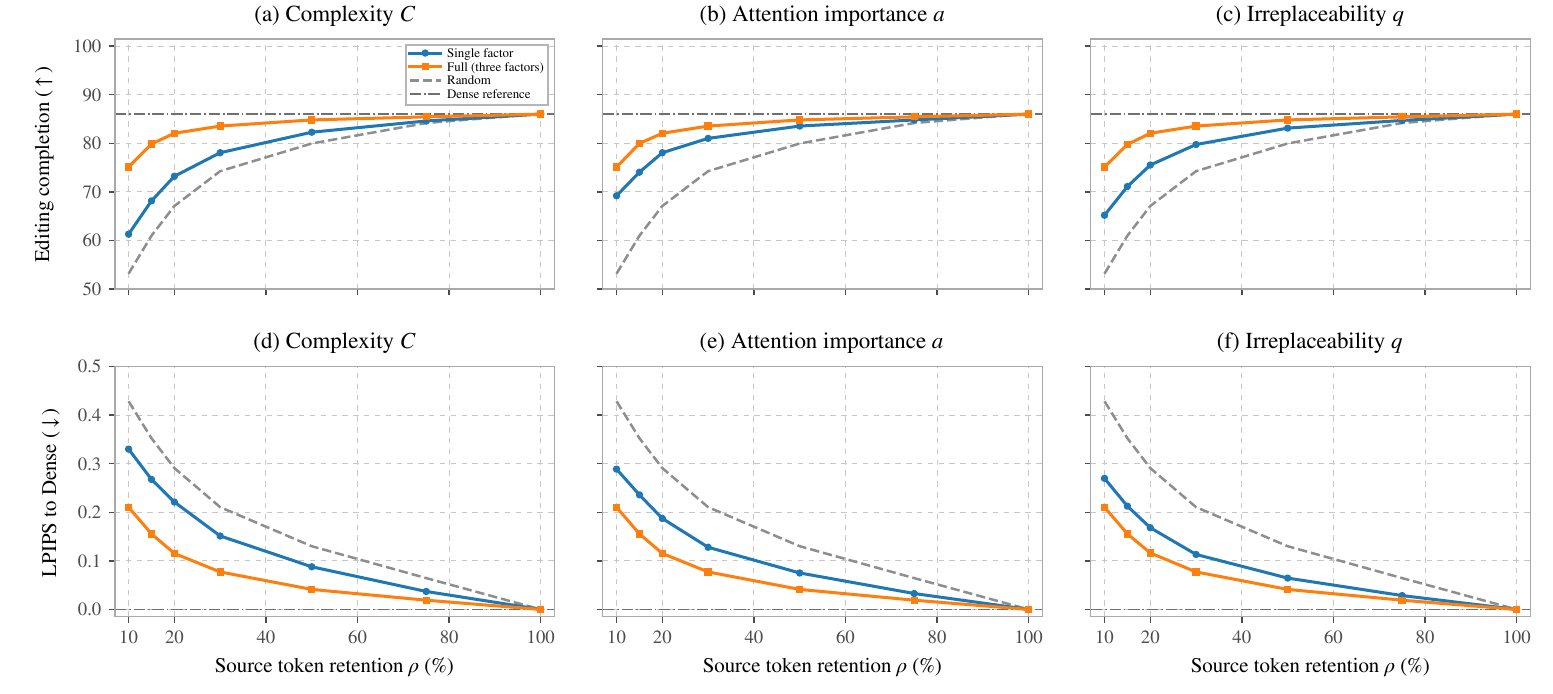}
    \caption{\textbf{Single-factor and joint selection
    across source token budgets.}
    Columns correspond to complexity, attention
    importance, and irreplaceability.
    The top row reports editing completion
    ($\uparrow$), and the bottom row reports LPIPS
    to Dense ($\downarrow$).
    Evaluation settings and factor analyses are
    detailed in Appendix~\ref{app:factors}.}
    \label{fig:retention_scan}
\end{figure}

\subsection{Does Demand-Weighting Improve Budget Utilization?}
\label{sec:demand_weighting}

\noindent\textbf{Weight comparison.}
Table~\ref{tab:weighting_effect} isolates the weighting
strategy at $20\%$ source retention, with compression,
maintenance, and external components held fixed.

Across three seeds, TriDiT improves video-level editing
completion over Uniform weighting by 1.65 points
($95\%$ CI $[1.10,2.20]$).
This cross-seed result is distinct from the single-seed
means in Table~\ref{tab:weighting_effect}.
Aggregation and statistical protocols are detailed in
Appendix~\ref{app:evaluation}; additional factor
analyses appear in Appendix~\ref{app:factors}.

\begin{table}[!ht]
    \centering
    \small
    \setlength{\tabcolsep}{3pt}

    \begin{tabular*}{\columnwidth}{
        @{\extracolsep{\fill}}lrrr@{}
    }
        \toprule
        Weighting
        & Edit $\uparrow$
        & Temp. $\uparrow$
        & DiT/s $\downarrow$ \\
        \midrule
        Uniform
        & 81.1 & 87.9 & 8.75 \\
        Attention-only
        & 81.6 & 88.5 & 8.70 \\
        Three-factor
        & \textbf{82.5}
        & \textbf{88.8}
        & \textbf{8.435} \\
        \bottomrule
    \end{tabular*}

    \caption{\textbf{Effect of demand weighting.}
    All variants retain 20\% of source tokens with
    Spark-H3 and SpotEdit enabled.
    Only the weighting strategy changes.
    Results use test seed 42.}
    \label{tab:weighting_effect}
\end{table}

\noindent\textbf{Retention-rate scans.}
Figure~\ref{fig:retention_scan} compares individual
factors, their joint strategy, and random selection
over source retention ratios from $10\%$ to $100\%$.

\setlength{\textfloatsep}{10pt}

\subsubsection{Multi-Budget and Efficiency Comparisons}
\label{sec:budget_efficiency}

Across the reported budgets, TriDiT improves editing
completion by $0.8$--$1.0$ points over ToMA-select,
the uniform-weight coverage control.
Equal retention ratios do not imply equal latency;
budget-wise comparisons are provided in
Appendix~\ref{app:baselines}.

\noindent\textbf{Budget efficiency at similar mean quality.}
Without external components, TriDiT at 20\% retention
achieves editing and temporal scores of 80.2 and 87.1,
compared with 80.1 and 87.0 for Attention-only at 30\%.
DiT latency decreases from $13.80$\,s to $12.229$\,s,
a $1.13\times$ speedup.
Similar means do not establish statistical equivalence
or a minimum feasible budget.
Detailed quality--latency comparisons and computational
analysis are provided in
Appendices~\ref{app:baselines} and~\ref{app:timing}.

\subsection{Why Retain Residuals and Perform Re-selection?}
\label{sec:maintenance_ablation}

Table~\ref{tab:maintenance_ablation} evaluates state
maintenance on the development set at $20\%$ source
retention, with Spark-H3 and SpotEdit enabled.
Removing mean aggregation, position residuals,
or periodic re-selection reduces editing scores
by 2.5, 3.5, and 3.0 points, respectively.
Recovery diagnostics, update-interval comparisons,
and representative re-entry analyses are provided in
Appendix~\ref{app:recovery}.

\begin{table}[!t]
    \centering
    \small
    \setlength{\tabcolsep}{4pt}
    \renewcommand{\arraystretch}{1.12}

    \begin{tabular*}{\linewidth}{
        @{\extracolsep{\fill}}lrrr@{}
    }
        \toprule
        Configuration
        & Edit $\uparrow$
        & Temp. $\uparrow$
        & DiT/s $\downarrow$ \\
        \midrule
        Full method
        & \textbf{82.5} & \textbf{88.8} & 8.7 \\
        No mean aggregation
        & 80.0 & 87.5 & 8.6 \\
        No position residuals
        & 79.0 & 85.8 & 8.6 \\
        \shortstack[l]{Initial selection only\\per denoising step}
        & 79.5 & 85.5 & \textbf{8.1} \\
        \bottomrule
    \end{tabular*}

    \caption{\textbf{Ablation of state maintenance.}
    Development-set results at 20\% source retention
    with Spark-H3 and SpotEdit enabled.
    The development baseline is $8.7$\,s.
    Additional analyses appear in
    Appendix~\ref{app:recovery}.}
    \label{tab:maintenance_ablation}
\end{table}

\input{sec/qualitative_figure}

\subsection{Overhead and Applicability}
\label{sec:overhead_applicability}

Cumulative maintenance events total $0.675$\,s,
approximately $8.00\%$ of the reported DiT latency.
Under overlapping execution, this event sum is not
the incremental wall-clock overhead.
Peak allocated GPU memory decreases by $12.5\%$,
from $47.2$\,GB to $41.3$\,GB.
Timing attribution and memory accounting are detailed
in Appendix~\ref{app:timing}.

Combined with Spark-H3, TriDiT achieves $4.30\times$
and $2.77\times$ speedups over Dense in high-resolution
and portrait scenarios.
Full-DiT speedups on UniVideo and UNIC are
$1.79\times$ and $1.85\times$, respectively;
the Wan-VACE condition-branch adaptation achieves
$1.15\times$.
Transfer configurations, compression scopes,
and corresponding quality results are provided in
Appendix~\ref{app:evaluation}.

\subsection{Annotated Qualitative Analysis}
\label{sec:qualitative}
Figure~\ref{fig:qualitative} examines details, context, and failure modes.

\noindent\textbf{Local details (A1).}
Jacket motifs and button boundaries remain visible after recoloring,
illustrating details that residual recovery aims to preserve
(Eq.~\eqref{eq:recovery}).

\noindent\textbf{Edited and retained regions (B1/B2).}
The mug becomes transparent while its blue neighbor remains recognizable;
altered handles and silhouettes indicate approximate geometric preservation.

\noindent\textbf{Occlusion (D1/D2).}
The rabbit retains its category and foreground relationship in selected
frames. Video time differs from network depth; this does not establish
re-selection's causal benefit.

\noindent\textbf{Limitations (C1/C2).}
Overlapping sheep have ambiguous boundaries, and grass texture changes.
Without paired Dense outputs, these errors cannot be attributed to compression.
Mechanism claims require matched ablations; further annotations appear in
Appendix~\ref{app:qual_local}.

\subsection{Scope of the Acceleration Claim}
\label{sec:interpretation}
The reported $3.61\times$ concerns DiT latency, not end-to-end inference.
For unchanged non-DiT cost $T_{\mathrm{fixed}}\ge0$ and
$S_{\mathrm{DiT}}\ge1$,
\begin{equation}
S_{\mathrm{end}}=
\frac{T_{\mathrm{fixed}}+T_{\mathrm{Dense}}}
{T_{\mathrm{fixed}}+T_{\mathrm{Dense}}/S_{\mathrm{DiT}}}
\le S_{\mathrm{DiT}}.
\label{eq:end_to_end_scope}
\end{equation}
This accounting relation is not an additional measured result.
Single-video latency and aggregate throughput require separate evaluation.

\noindent\textbf{Reading quality and efficiency together.}
GT metrics assess edited-reference quality; Dense metrics quantify
approximation fidelity. High Dense similarity may reproduce reference
errors, while SSIM-S separately evaluates source preservation.
The combined system reduces latency from $12.229$\,s to $8.435$\,s,
while Dense LPIPS increases from $0.039$ to $0.055$.
Individual speedups should not be multiplied because the components
affect overlapping computation.

\noindent\textbf{Recovery trade-offs.}
Residuals retain positional information but cannot reproduce omitted
nonlinear updates. Re-selection lets merged positions compete for capacity
again, without guaranteeing correction of accumulated errors.
The ablations support both mechanisms in the tested setting.
Shorter layer groups permit more frequent adaptation; longer groups
amortize maintenance costs. The update interval therefore trades
adaptation frequency against execution overhead.

%% file: sec/qualitative_figure.tex
\begin{figure*}[!t]
\centering
\begin{minipage}[t]{0.485\textwidth}
\centering
{\small\textbf{(a) Attribute editing}}\\[1pt]
{\small To \textcolor{qualblue}{\textit{``a blue jacket''}}}\\[3pt]
\begin{tabular}{@{}c@{\hspace{3pt}}c@{}}
\small Source & \small TriDiT \\[2pt]
\ROIImage[viewport=150 128 620 478,clip,width=0.493\linewidth]{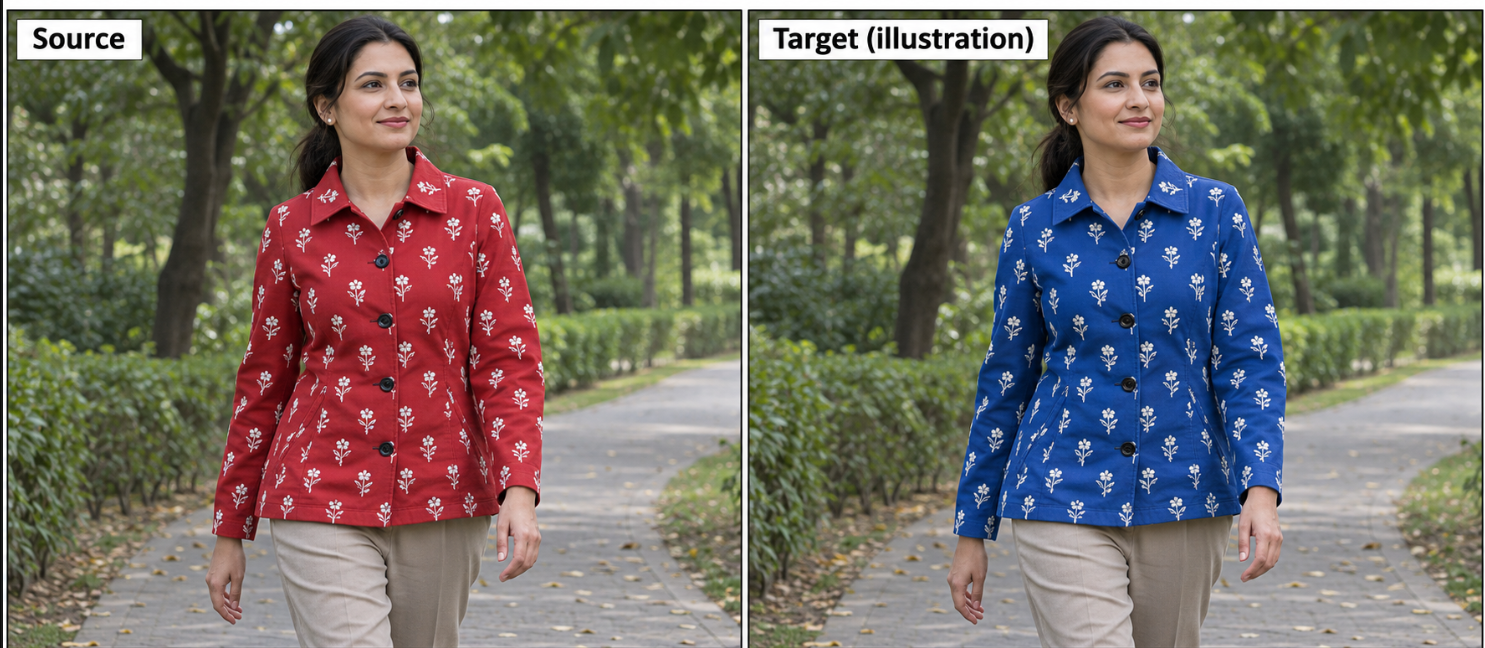}{%
\Roi{qualblue}{(.277,.414)}{(.702,.629)}{A1}} &
\ROIImage[viewport=895 128 1365 478,clip,width=0.493\linewidth]{Figure/qual_clothing.png}{%
\Roi{qualblue}{(.277,.414)}{(.702,.629)}{A1}} \\[3pt]
\includegraphics[viewport=280 273 480 348,clip,width=0.493\linewidth]{Figure/qual_clothing.png} &
\includegraphics[viewport=1025 273 1225 348,clip,width=0.493\linewidth]{Figure/qual_clothing.png} \\[2pt]
\multicolumn{2}{c}{\scriptsize\textbf{A1:} texture detail $\vert$ residual recovery}
\end{tabular}
\end{minipage}\hfill
\begin{minipage}[t]{0.485\textwidth}
\centering
{\small\textbf{(b) Material editing}}\\[1pt]
{\small To \textcolor{qualblue}{\textit{``a transparent glass mug''}}}\\[3pt]
\begin{tabular}{@{}c@{\hspace{3pt}}c@{}}
\small Source & \small TriDiT \\[2pt]
\ROIImage[viewport=10 5 1017 755,clip,width=0.493\linewidth]{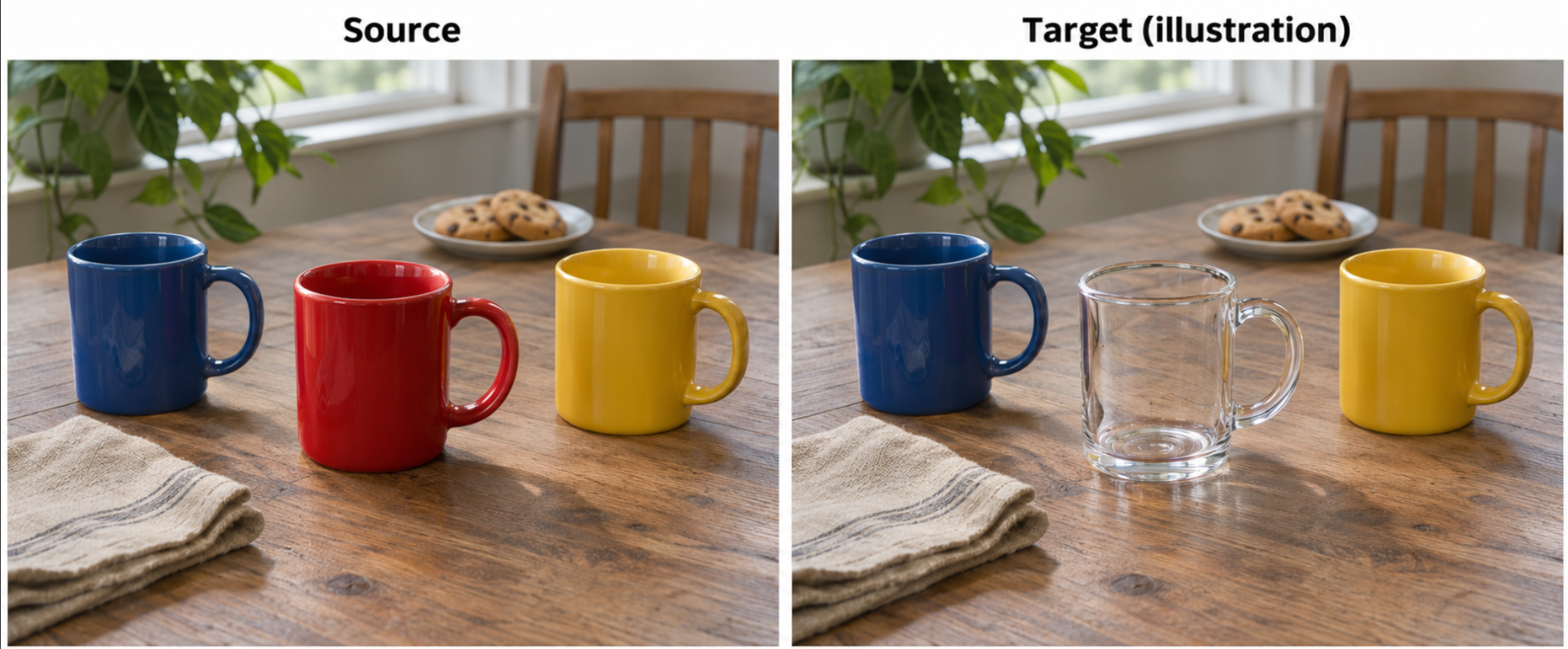}{%
\Roi{qualblue}{(.372,.457)}{(.665,.604)}{B1}
\Roi{qualgreen}{(.07,.40)}{(.33,.71)}{B2}} &
\ROIImage[viewport=1036 5 2040 753,clip,width=0.493\linewidth]{Figure/qual_material.png}{%
\Roi{qualblue}{(.373,.458)}{(.666,.606)}{B1}
\Roi{qualgreen}{(.07,.40)}{(.33,.71)}{B2}} \\[3pt]
\includegraphics[viewport=385 348 680 458,clip,width=0.493\linewidth]{Figure/qual_material.png} &
\includegraphics[viewport=1410 348 1705 458,clip,width=0.493\linewidth]{Figure/qual_material.png} \\[2pt]
\multicolumn{2}{c}{\scriptsize\textbf{B1/B2:} target/context $\vert$ demand weighting}
\end{tabular}
\end{minipage}
\par\vspace{6pt}
\begin{minipage}[t]{0.485\textwidth}
\centering
{\small\textbf{(c) Multi-object editing}}\\[1pt]
{\small To \textcolor{qualblue}{\textit{``a flock of sheep''}}}\\[3pt]
\begin{tabular}{@{}c@{\hspace{3pt}}c@{}}
\small Source & \small TriDiT \\[2pt]
\ROIImage[viewport=6.75 15 403.45 223.47,clip,width=0.493\linewidth]{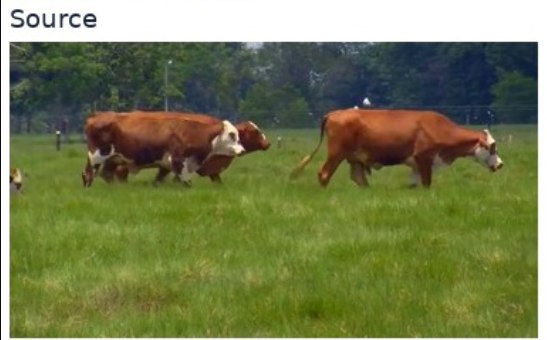}{%
\Roi{qualorange}{(.106,.475)}{(.509,.691)}{C1}
\Roi{qualorange}{(.52,.09)}{(.78,.30)}{C2}} &
\ROIImage[viewport=32 25 1243 662,clip,width=0.493\linewidth]{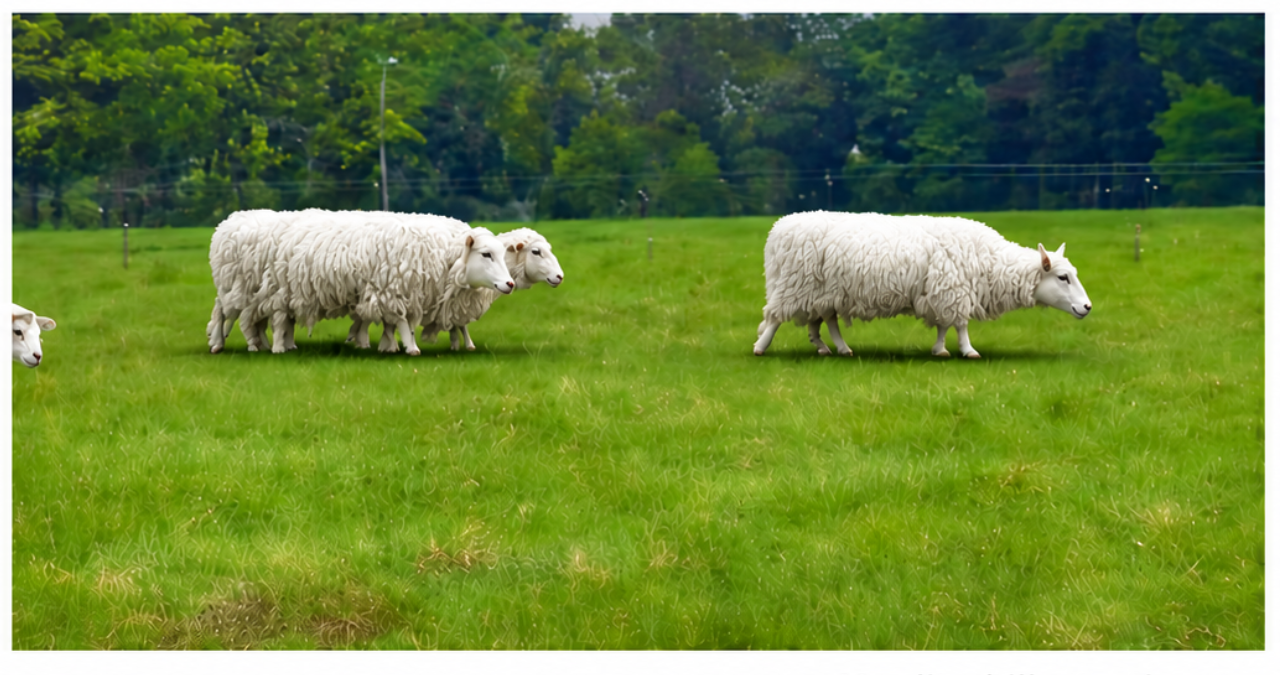}{%
\Roi{qualorange}{(.126,.455)}{(.457,.632)}{C1}
\Roi{qualorange}{(.52,.09)}{(.78,.30)}{C2}} \\[3pt]
\includegraphics[viewport=48.74 113.99 208.47 158.98,clip,width=0.493\linewidth]{Figure/qual_cows.png} &
\includegraphics[viewport=185 315 585 427.5,clip,width=0.493\linewidth]{Figure/qual_sheep.png} \\[2pt]
\multicolumn{2}{c}{\scriptsize\textbf{C1/C2:} overlap and background changes}
\end{tabular}
\end{minipage}\hfill
\begin{minipage}[t]{0.485\textwidth}
\centering
{\small\textbf{(d) Editing across occlusion}}\\[1pt]
{\small To \textcolor{qualblue}{\textit{``a white rabbit''}}}\\[3pt]
\begin{tabular}{@{}c@{\hspace{2pt}}c@{\hspace{2pt}}c@{\hspace{2pt}}c@{}}
& \small Before & \small Occluded & \small Reappeared \\[2pt]
\raisebox{14pt}{\rotatebox{90}{\scriptsize Source}} &
\includegraphics[viewport=138 302 567 582,clip,width=0.309\linewidth]{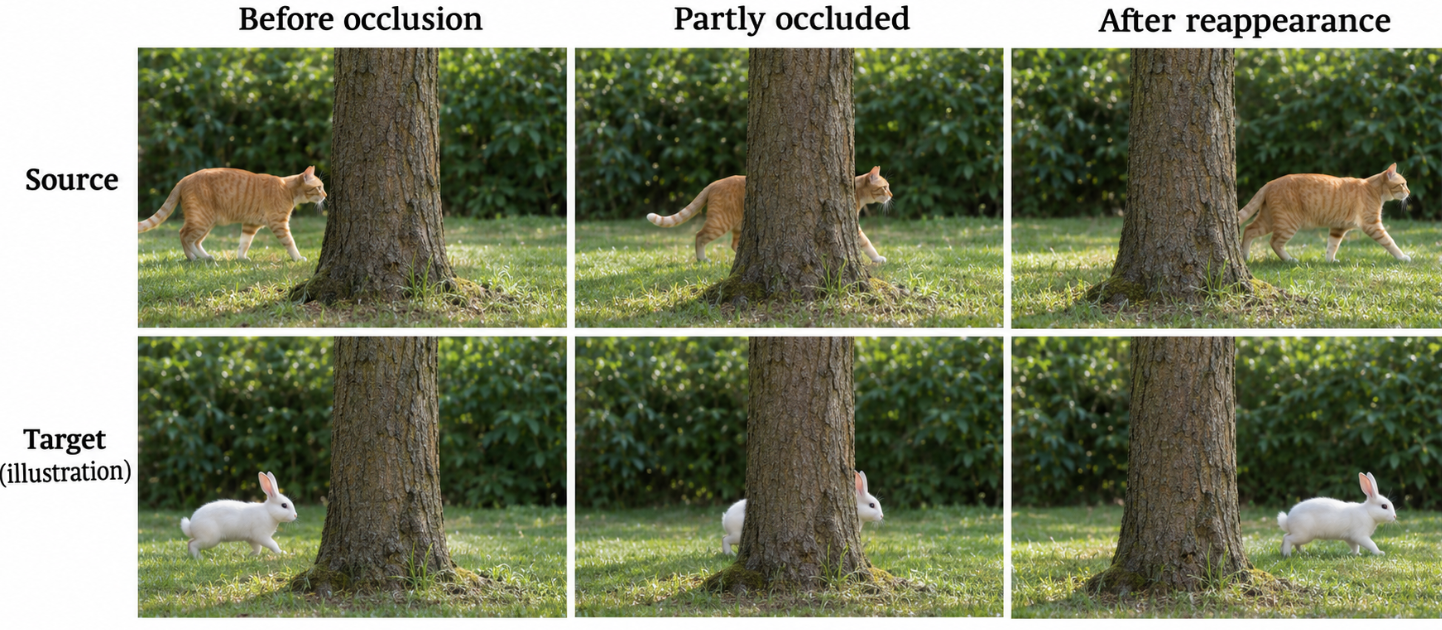} &
\includegraphics[viewport=575 302 1003 582,clip,width=0.309\linewidth]{Figure/qual_occlusion.png} &
\includegraphics[viewport=1011 302 1441 582,clip,width=0.309\linewidth]{Figure/qual_occlusion.png} \\[1pt]
\raisebox{14pt}{\rotatebox{90}{\scriptsize TriDiT}} &
\ROIImage[viewport=138 12 567 293,clip,width=0.309\linewidth]{Figure/qual_occlusion.png}{%
\Roi{qualgreen}{(.08,.20)}{(.40,.54)}{D2}} &
\ROIImage[viewport=575 12 1003 293,clip,width=0.309\linewidth]{Figure/qual_occlusion.png}{%
\Roi{qualblue}{(.63,.18)}{(.77,.55)}{D1}} &
\ROIImage[viewport=1011 12 1441 293,clip,width=0.309\linewidth]{Figure/qual_occlusion.png}{%
\Roi{qualgreen}{(.60,.20)}{(.90,.54)}{D2}} \\[2pt]
\multicolumn{4}{c}{\scriptsize\textbf{D1/D2:} occlusion/reappearance $\vert$ re-selection}
\end{tabular}
\end{minipage}
\caption{\textbf{Qualitative analysis with aligned regional comparisons.}
In (a)--(c), the lower row enlarges the boxed A1, B1, and C1 regions;
(d) compares selected times before, during, and after occlusion.
Blue/green boxes mark edited details or retained context; orange boxes
mark visible limitations. Mechanism labels indicate diagnostic questions,
not causal ablation findings or internal-state measurements.}
\label{fig:qualitative}
\end{figure*}

%% file: sec/5_conclusion.tex
\section{Conclusion and Limitations}
\label{sec:conclusion}

\noindent\textbf{Conclusion.}
We presented TriDiT, combining demand-weighted source-token selection, compressed execution, and residual-based token unmerging for video DiT editing. Under a shared seven-step setup, TriDiT achieves a $2.49\times$ DiT speedup with small quality losses. With Spark-H3 and SpotEdit, it reaches $3.61\times$ over Dense and $1.75\times$ over their combination.

\par\smallskip
\noindent\textbf{Limitations.}
Weighted coverage is a proxy objective, and fixed intra-group residuals prevent exact recovery of Dense trajectories. The offline-trained predictor faces distribution shifts on restored states. Temporal gains over attention-only selection remain modest and statistically unconfirmed. Speedups depend on residual storage and backbone interfaces, requiring broader validation.
Long videos and rapid scene changes warrant further evaluation of
representative coverage and accumulated reconstruction errors.

%% file: sec/supplement.tex
\clearpage
\appendix
\setcounter{figure}{0}\renewcommand{\thefigure}{S\arabic{figure}}
\setcounter{table}{0}\renewcommand{\thetable}{S\arabic{table}}
\setcounter{equation}{0}\renewcommand{\theequation}{S\arabic{equation}}
\renewcommand{\theHfigure}{supp.\arabic{figure}}
\renewcommand{\theHtable}{supp.\arabic{table}}
\renewcommand{\theHequation}{supp.\arabic{equation}}
\newcommand{\suppfig}[1]{%
  \begin{figure}[H]\centering
  \includegraphics[page=#1,width=\linewidth,height=.24\textheight,keepaspectratio]{supplement_figures/panels.pdf}
  \caption[Supplementary figure #1]{\input{supplement_figures/caption_s#1.tex}}\label{fig:s#1}
  \end{figure}}
\renewcommand{\theHsection}{supp.\Alph{section}}
\setlength{\intextsep}{5pt}
\section*{Supplementary Material}
\section{Predictor and implementation details}\label{app:implementation}
\subsection{Configuration}
This supplement follows Sections~3 and~4 of the main paper. Only the attention predictor is trained. The DiT, text encoder, VAE and merged distillation LoRA remain frozen. Table~\ref{tab:s1} gives the operating point for the main H3 evaluation. The qualitative comparisons identify separate four-step runs where applicable.
\begin{table}[H]\centering\small
\begin{tabular}{@{}p{.23\linewidth}p{.71\linewidth}@{}}\toprule
Setting & Configuration\\\midrule
Hardware & Four A100 80GB GPUs per video, layer partitioning and FP16 mixed precision\\
Input & $1344\times768$, 81 frames at 24 fps\\
Visual tokens & $21\times42\times24$ positions per branch, $N_s=N_t=21{,}168$\\
Backbone & 50 layers, hidden width $5{,}376$; 56 attention heads of width 128 (attention width $7{,}168$)\\
Sampling & Seven denoising steps, merged LoRA, CFG disabled\\
Source budget & $K=\lfloor0.2N_s\rfloor=4{,}233$, refresh every five layers\\
Predictor & Eight query groups, projection width 32, 921,088 parameters\\
Training & 16 boundary states per update, 20,000 updates, 500-step warm-up and cosine decay\\
Optimizer & AdamW, learning rate $10^{-4}$, weight decay 0.01\\\bottomrule
\end{tabular}
\caption{Training and inference configuration. A boundary state is a predictor-training sample. Hidden width and attention width are distinct configuration dimensions.}\label{tab:s1}
\end{table}
\subsection{Predictor supervision}
Training uses dense trajectories from all seven steps and all representative-update boundaries on 800 editing tasks. At a boundary, the teacher uses the next layer's normalization and positional encoding. For each head, divide target queries into eight contiguous groups. For each group, retain the source columns of teacher attention, take the maximum over valid queries, and normalize over source positions.

\noindent\textbf{Projections.} For each group, concatenate the element-wise mean, maximum and minimum of its target queries. Separate affine maps project this descriptor and the source keys into $d'=32$ dimensions. Heads have independent parameters shared across layers and denoising steps. With $H=56$ and $d_h=128$, the predictor parameter count is
\begin{equation}
H\big[(3d_h+1)+(d_h+1)\big]d'=921{,}088.
\end{equation}
For head $h$ and group $g$, the predicted distribution is the softmax over source positions of the projected query--key products scaled by $1/\sqrt{d'}$. Writing $A^h_{qi}$ for the teacher's attention and $z_g^h,k_i^{\prime h}$ for the two projected descriptors, the distributions are
\begin{equation}
\begin{aligned}
p^h_{gi}&=\frac{\max_{q\in g}A^h_{qi}}{\sum_{j\in\mathcal S}\max_{q\in g}A^h_{qj}},\\
\widehat p^h_{gi}&=\operatorname{softmax}_{i\in\mathcal S}\left(\frac{(z_g^h)^\top k_i^{\prime h}}{\sqrt{d'}}\right).
\end{aligned}
\end{equation}
The loss weights each group's teacher-to-predictor KL divergence by its share of valid queries. Padding is excluded. The predictor is trained on detached dense states, without auxiliary losses or temperature scaling.

\noindent\textbf{Execution.} At inference, average the predicted demand across heads, weighting groups by valid query count. Within each denoising step, selection takes place at boundaries $0,5,\ldots,45$. States are restored before the next selection.
\subsection{State storage and component interfaces}
Each group stores $K$ representatives, one residual vector and one representative assignment per source position. Source tokens retain their original spatiotemporal coordinates. Spark-H3 sparsifies interactions on the resulting sequence. SpotEdit updates active targets and reuses cached inactive states while preserving the logical target layout and key/value context. The downstream interface determines whether the final source states must be restored.
\subsection{Coverage and implementation}
The weights and sparse graph are fixed for one selection round. Exact lazy greedy accepts a recomputed marginal gain once it dominates the remaining valid upper bounds. Select exactly $K$ representatives and resolve ties by source index. The coverage guarantee applies within each round. Complexity is computed once using valid four-neighbors at spacings 1 and 2. Retrieval considers $6\times6$ blocks, adjacent blocks including diagonals, and neighboring time slices. Reranking uses cosine similarity in the original state space.

\section{Baseline adaptations and budget efficiency}\label{app:baselines}
\subsection{Controlled comparisons}
All standalone comparisons use a full target branch. Attention-only and ToMA-select share the TriDiT execution pipeline. Full ToMA retains its selection and merging procedure.

At 20\% retention, TriDiT reaches 27.63 dB GT PSNR, 0.41 dB above Full ToMA and 0.21 dB below the uncompressed reference. LPIPS to the reference is 0.039 for TriDiT and 0.054 for Full ToMA.
\begin{table}[H]\centering\small
\begin{tabular}{@{}p{.39\linewidth}p{.55\linewidth}@{}}\toprule
Method & Selection or execution\\\midrule
MiniMax-H3-ref2VA & Full source tokens, seven-step reference\\
Attention-only & Attention demand only\\
ToMA-select & Uniform demand\\
TriDiT & Complexity, attention and irreplaceability\\
Full ToMA & Complete selection and merging adaptation\\\bottomrule
\end{tabular}
\caption{Variants used in the controlled comparisons.}\label{tab:s2}
\end{table}
\begin{figure}[H]\centering
\includegraphics[page=01,width=.97\linewidth]{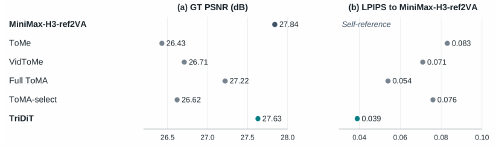}
\caption[Supplementary figure 01]{\input{supplement_figures/caption_s01.tex}}\label{fig:s01}
\end{figure}
\subsection{Quality across budgets}
\begin{figure}[H]\centering
\includegraphics[page=02,width=.97\linewidth]{supplement_figures/panels.pdf}
\caption[Supplementary figure 02]{\input{supplement_figures/caption_s02.tex}}\label{fig:s02}
\end{figure}
At 20\%, TriDiT and ToMA-select score 80.2 and 79.3 on editing and 87.1 and 86.4 on temporal consistency. Across the five compressed budgets, the editing difference is 0.8--1.0 points.

\noindent\textbf{Matched-quality operating points.} Attention-only at 30\% and TriDiT at 20\% differ by 0.1 points on both reported means. Figure~\ref{fig:s03} shows their DiT times. The comparison describes these operating points without establishing statistical equivalence.
\begin{figure}[H]\centering
\includegraphics[page=03,width=.97\linewidth]{supplement_figures/panels.pdf}
\caption[Supplementary figure 03]{\input{supplement_figures/caption_s03.tex}}\label{fig:s03}
\end{figure}

\section{Residual recovery and re-selection}\label{app:recovery}
\subsection{Incremental recovery}
Let $G_u$ denote the set assigned to representative $u$, $h_i$ the entry state of token $i$, and $m(i)$ its assignment. Mean aggregation and the stored residual satisfy
\begin{equation}
\bar h_u=|G_u|^{-1}\sum_{i\in G_u}h_i,\qquad e_i=h_i-\bar h_{m(i)}.
\end{equation}
After the layer group updates the representative to $\bar h'_u$, recovery gives
\begin{equation}
\widetilde h'_i=\bar h'_{m(i)}+e_i=h_i+(\bar h'_{m(i)}-\bar h_{m(i)}).
\end{equation}
Tokens with the same assignment therefore preserve their entry-state differences. Recovery approximates the uncompressed states and makes every source position eligible for re-selection.
\subsection{Maintenance ablation}
\begin{figure}[H]\centering
\includegraphics[page=04,width=.97\linewidth]{supplement_figures/panels.pdf}
\caption[Supplementary figure 04]{\input{supplement_figures/caption_s04.tex}}\label{fig:s04}
\end{figure}
This ablation uses 20\% retention with Spark-H3 and SpotEdit. Its development latency is 8.7 s. The separate test timing is 8.435 s (Table~\ref{tab:s6}).
\subsection{Recovery and selection diagnostics}
A representative can change when a position is reconsidered after restoration. An assignment trace at successive group boundaries records this change for each source position.

\section{Demand-factor analysis}\label{app:factors}
\subsection{Factor controls}
All controls use the same graph, aggregation, recovery and representative schedule. The three-factor demand weights are given in Table~\ref{tab:s3}.
\begin{table}[H]\centering\small
\begin{tabular}{@{}lcl@{}}\toprule
Signal & Weight & Refresh\\\midrule
Attention demand & 0.40 & Each boundary\\
Irreplaceability & 0.35 & Each boundary\\
Complexity & 0.25 & Once per generation\\\bottomrule
\end{tabular}
\caption{Demand factors in the representative-selection score.}\label{tab:s3}
\end{table}
\begin{figure}[H]\centering
\includegraphics[page=05,width=.97\linewidth]{supplement_figures/panels.pdf}
\caption[Supplementary figure 05]{\input{supplement_figures/caption_s05.tex}}\label{fig:s05}
\end{figure}
\subsection{Paired comparisons}
The uniform control scores 81.1 on editing and 87.9 on temporal consistency at seed 42. Across three seeds, the mean paired editing gain over uniform selection is 1.65 points (95\% confidence interval 1.10 to 2.20). Displayed three-seed means differ by approximately 0.9 points from attention-only selection. The seed-42 paired temporal difference against attention-only is 0.30 points (95\% confidence interval $-0.22$ to 0.82). This interval includes zero. Paired comparisons aggregate tasks within source video.

\section{Timing, memory and computational cost}\label{app:timing}
\subsection{Measurement scope}
Table~\ref{tab:s6} reports end-to-end, attention-core and DiT latency separately. End-to-end timing covers input processing, encoding, denoising, decoding and output. The attention core covers QK, softmax and AV within the DiT. Four A100 80GB GPUs jointly generate one video. The H3 efficiency subset comprises 50 matched tasks at seed 42, using seven denoising steps and disabled CFG. The DiT interval begins before method-specific maintenance and ends after DiT output and cache updates. It excludes sampler updates, common text/VAE processing and I/O.
\subsection{Aggregation and attribution}
One run per task contributes to the efficiency mean. Speedup divides the mean reference latency by the mean method latency over the paired task set. Five fixed-task repeats assess repeatability separately.
\begin{figure}[H]\centering
\includegraphics[page=06,width=.97\linewidth]{supplement_figures/panels.pdf}
\caption[Supplementary figure 06]{\input{supplement_figures/caption_s06.tex}}\label{fig:s06}
\end{figure}
The standalone method changes editing and temporal scores by $-1.0$ and $-1.4$ points relative to the reference. The full combination gains 1.5 editing points over Spark-H3 + SpotEdit.

\noindent\textbf{Memory and maintenance.} The main paper reports peak allocated memory of 47.2 and 41.3 GB. The available summary does not specify device aggregation, byte conversion or warm-up count for these peaks. Maintenance events sum to 0.675 s (8.00\% of 8.435 s) and may overlap computation. Representatives, residuals and assignments occupy $O(KD)$, $O(N_sD)$ and $O(N_s)$ storage, respectively.

\section{Evaluation protocols and transfer}\label{app:evaluation}
\subsection{Data and evaluation sets}
Source videos are disjoint across splits, with four instructions per video. Training uses 200 videos (800 tasks), development 50 (200 tasks) and testing 100 (400 tasks). The test set contains 380 non-style edits and 20 stylizations. The efficiency subset samples 40 non-style edits and 10 stylizations without replacement from the test categories. Matched H3 test runs share inputs, seed 42 and initial noise.
\subsection{References and metrics}\label{app:metrics}
\begin{table}[H]\centering\small
\begin{tabular}{@{}p{.33\linewidth}p{.61\linewidth}@{}}\toprule
Metric & Reference and support\\\midrule
SSIM-S & Source pixels in valid unedited windows, 380 tasks\\
GT PSNR/SSIM/LPIPS & Edited Ditto-1M reference, full valid image, 400 tasks\\
GT FID & 16 frames per video, 6,400 frames per set\\
Masked fidelity & Matching backbone output in non-edited regions, 380 tasks\\
Full-frame fidelity & Matching backbone output, 400 tasks, 81 frames per task\\\bottomrule
\end{tabular}
\caption{Evaluation references and spatial support.}\label{tab:s4}
\end{table}
\noindent\textbf{Masks.} One annotator corrects SAM 2.1 Hiera-Large masks and a second reviews them. All methods use the same masks. SSIM-S uses RGB in $[0,1]$ and $11\times11$ Gaussian windows ($\sigma=1.5$) wholly inside valid unedited regions. Average windows within tasks, then average across tasks.

\noindent\textbf{Full-frame metrics.} Edited Ditto-1M references are manually screened against the source. Align frame indices, apply identical resizing and cropping, and exclude padding. Tasks receive equal weight. LPIPS uses AlexNet v0.1 with RGB in $[-1,1]$. Clean-FID uses Inception-v3 on 16 evenly spaced frames per clip, including both endpoints. For H3 fidelity, match prompt, weights, noise and sampler; evaluate all 32,400 frame pairs without registration and average PSNR in dB.
\subsection{Human ratings}
Three independent raters view the source, instruction and output in randomized order, with the method and runtime hidden. Editing anchors 1, 3 and 5 denote failed, partial and sustained edits. Temporal anchors denote severe persistent defects, clear local defects and no obvious defects. Intermediate scores interpolate the anchors. Raters exclude requested changes and normal movement from temporal defects. Average the three ratings per task and rescale from $[1,5]$ to $[0,100]$. With task $n$, rater $j$ and metric $k$, the reported score over $N$ tasks is
\begin{equation}
s_k=\frac{25}{N}\sum_{n=1}^{N}\left(\frac{1}{3}\sum_{j=1}^{3}r_{njk}-1\right).
\end{equation}
For three-seed analyses (42, 123 and 2024), aggregate tasks and seeds within each of the 100 source videos before computing paired differences. Individual ratings are needed to quantify inter-rater agreement.
\subsection{Transfer across backbones}
\begin{table}[H]\centering\small
\begin{tabular}{@{}lccc@{}}\toprule
Backbone & DiT index & Speedup & Attn. index\\\midrule
UniVideo & 56 & $1.79\times$ & 39\\
UNIC & 54 & $1.85\times$ & 38\\
Wan-VACE & 87 & $1.15\times$ & 81\\\bottomrule
\end{tabular}
\caption{Transfer at 20\% retention. Each uncompressed backbone has index 100 within each timing scope.}\label{tab:s5}
\end{table}
\begin{figure}[H]\centering
\includegraphics[page=07,width=.97\linewidth]{supplement_figures/panels.pdf}
\caption[Supplementary figure 07]{\input{supplement_figures/caption_s07.tex}}\label{fig:s07}
\end{figure}
UniVideo compresses source-video tokens, UNIC joint-sequence reference tokens, and Wan-VACE context-adapter states. The normalized indices refer to the series in main-paper Section~4.5. Paired visual comparisons appear in Sec.~F.6. Separate high-resolution and portrait evaluations with Spark-H3 report $4.30\times$ and $2.77\times$ DiT speedups.
\subsection{Per-video latency and quality}
\begin{table}[H]\centering\small
\begin{tabular}{@{}lrrr@{}}\toprule
Method & E2E & Attn. & DiT\\\midrule
MiniMax-H3-ref2VA & 35.5 & 17.4 & 30.45\\
InsV2V & 46.7 & 23.8 & 41.70\\
Lucy-Edit-dev & 29.5 & 14.0 & 24.50\\
Ditto & 28.4 & 13.3 & 23.40\\
VACE & 37.8 & 18.7 & 32.80\\
ToMe adaptation & 16.9 & 6.8 & 11.85\\
VidToMe adaptation & 17.1 & 6.9 & 12.05\\
Full ToMA adaptation & 17.6 & 7.2 & 12.60\\
ToMA-select & 17.1 & 6.9 & 12.10\\
TriDiT (standalone) & 17.2 & 7.00* & 12.229\\
TriDiT + Spark-H3 & & &\\
\quad + SpotEdit & 13.4 & 4.8 & 8.435\\\bottomrule
\end{tabular}
\caption{Mean per-video latency (seconds) on four A100 80GB GPUs. *Seven-step attention-core time covers QK, softmax and AV.}\label{tab:s6}
\end{table}
\begin{figure}[H]\centering
\includegraphics[page=08,width=.97\linewidth]{supplement_figures/panels.pdf}
\caption[Supplementary figure 08]{\input{supplement_figures/caption_s08.tex}}\label{fig:s08}
\end{figure}
The full system takes 13.4 s per video, 20.7\% less time than ToMe (16.9 s), the fastest listed baseline. Relative to MiniMax-H3-ref2VA, the end-to-end speedup is $35.5/13.4=2.65\times$ and the DiT-stage speedup is $30.45/8.435=3.61\times$. Standalone TriDiT takes 17.2 s at 27.63 dB. The full system saves another 3.8 s at 27.41 dB.
\input{sec/supplement_qualitative}

%% file: supplement_figures/caption_s01.tex
GT PSNR and LPIPS to the matching MiniMax-H3-ref2VA output (main-paper Table 1). Standalone adaptations retain 20\% of source tokens.

%% file: supplement_figures/caption_s02.tex
Editing completion and temporal consistency across six budgets (400 test tasks, seed 42). Both curves share the uncompressed endpoint.

%% file: supplement_figures/caption_s03.tex
Standalone DiT latency and editing and temporal scores for the matched-quality comparison.

%% file: supplement_figures/caption_s04.tex
Removing aggregation, residual recovery or periodic selection. Quality differences use the full development configuration as reference (editing 82.5, temporal 88.8).

%% file: supplement_figures/caption_s05.tex
Demand weighting at 20\% retention, with Spark-H3 and SpotEdit enabled. Quality changes are relative to the uniform control; latency is absolute.

%% file: supplement_figures/caption_s06.tex
Seven-step DiT latency relative to the 30.45 s MiniMax-H3-ref2VA reference. The full system is 1.75$\times$ faster than Spark-H3 + SpotEdit alone on this stage.

%% file: supplement_figures/caption_s07.tex
Editing and temporal score changes relative to the corresponding uncompressed backbone in the main-paper transfer experiment.

%% file: supplement_figures/caption_s08.tex
GT PSNR versus end-to-end time. Quality uses the edited reference from main-paper Table 1. Each point represents one configuration.

%% file: sec/supplement_qualitative.tex
\clearpage\onecolumn
\small
\setlength{\intextsep}{4pt}
\captionsetup{font=small,skip=3pt}
\renewcommand{\suppfig}[1]{%
\begin{figure}[H]\centering
\includegraphics[page=#1,width=.97\linewidth]{supplement_figures/panels.pdf}
\caption[Supplementary figure #1]{\input{supplement_figures/caption_s#1.tex}}\label{fig:s#1}
\end{figure}}

\subsection{Qualitative transfer across backbones}\label{app:transfer_visual}
Source frames correspond to Cases 14, 16 and 08. Workload: four A100 GPUs, 107 frames at $832\times480$, 50 steps and 20\% retention.

\noindent\textbf{UniVideo (Case 14).} Replace only the running tan dog with a white wolf of similar size. Preserve the dog's original run, jump, landing and timing. Keep the crouching person and snowy surroundings unchanged.
\begin{figure}[H]\centering
\includegraphics[page=09,width=.97\linewidth]{supplement_figures/panels.pdf}
\caption[Supplementary figure 09]{\input{supplement_figures/caption_s09.tex}}\label{fig:s09}
\end{figure}
\noindent\textbf{UNIC (Case 16).} Change only the light-colored car closest to the center of the intersection to bright red. Preserve the car's shape and trajectory, all other vehicles, road markings and buildings.
\begin{figure}[H]\centering
\includegraphics[page=10,width=.97\linewidth]{supplement_figures/panels.pdf}
\caption[Supplementary figure 10]{\input{supplement_figures/caption_s10.tex}}\label{fig:s10}
\end{figure}
\noindent\textbf{Wan-VACE (Case 08).} Replace the waterfall and rocks behind the woman with a snowy pine forest. Preserve the woman, her hair, face, clothing, pose and movement exactly. Keep the video monochrome.
\begin{figure}[H]\centering
\includegraphics[page=11,width=.97\linewidth]{supplement_figures/panels.pdf}
\caption[Supplementary figure 11]{\input{supplement_figures/caption_s11.tex}}\label{fig:s11}
\end{figure}

\clearpage
\section{Qualitative analysis}\label{app:qualitative}
\noindent\textbf{Three complementary criteria.} We inspect (i) editing completion: whether the requested addition, removal or appearance change is visible; (ii) temporal consistency: whether the edited appearance and identity persist across the sampled times without obvious disappearance or deformation; and (iii) non-edited-region preservation: whether surrounding people, objects and scene geometry remain consistent with the source at matched times. Motion preservation compares the source and edited action, whereas temporal consistency concerns continuity within the output. Transparency is assessed as part of material-edit completion. For global stylization, preservation concerns structure and motion rather than unchanged color or texture. The sampled frames support qualitative observations, not a guarantee of flicker-free playback or a measured success rate.

\noindent\textbf{Quality and efficiency.} The examples identify which visible properties TriDiT retains under compression. Table~\ref{tab:s6} separately reports a $2.06\times$ mean end-to-end speedup ($35.5/17.2$) and $2.49\times$ DiT speedup ($30.45/12.229$) over MiniMax-H3-ref2VA for standalone TriDiT. These benchmark-level gains contextualize comparable qualitative outcomes; they are not measurements of the individual illustrated clips.

The rightmost columns, labeled E2E, report per-video end-to-end latency in seconds. Table~\ref{tab:s6} provides the model-level means. TriDiT uses the standalone configuration. Source rows have no editing latency.
\subsection{Local edits and comparisons with public baselines}\label{app:qual_local}
\noindent\textbf{Selective removal in a walking group.}\par
\noindent\href{https://zhw-zhang.github.io/ReCo-page/static/videos/comparison/remove_ori.mp4}{ReCo.} Remove the woman with glasses on the left.
\begin{figure}[H]\centering
\includegraphics[page=12,width=.97\linewidth]{supplement_figures/panels.pdf}
\caption[Supplementary figure 12]{\input{supplement_figures/caption_s12.tex}}\label{fig:s12}
\end{figure}

\clearpage
\noindent\textbf{Public object-removal and addition examples.}\par
\noindent\href{https://viva-paper.github.io/assets/comparison/ins/083/src.mp4}{VIVA 083.} Remove the bowl in the middle.
\begin{figure}[H]\centering
\includegraphics[page=13,width=.97\linewidth]{supplement_figures/panels.pdf}
\caption[Supplementary figure 13]{\input{supplement_figures/caption_s13.tex}}\label{fig:s13}
\end{figure}
\noindent\href{https://viva-paper.github.io/assets/comparison/ins/064/src.mp4}{VIVA 064.} Add a white dog beside the woman.
\begin{figure}[H]\centering
\includegraphics[page=14,width=.97\linewidth]{supplement_figures/panels.pdf}
\caption[Supplementary figure 14]{\input{supplement_figures/caption_s14.tex}}\label{fig:s14}
\end{figure}

\clearpage
\subsection{Boundary preservation and motion}
\noindent\textbf{Case 13: Jacket boundary at the wrist.} Change only the beige jacket worn by the curly-haired woman on the left to bright red. Preserve the clothing of the other two people, their faces, hands and the table.
\begin{figure}[H]\centering
\includegraphics[page=15,width=.97\linewidth]{supplement_figures/panels.pdf}
\caption[Supplementary figure 15]{\input{supplement_figures/caption_s15.tex}}\label{fig:s15}
\end{figure}
\noindent\textbf{Case 14: Running dog to white wolf.} Replace only the running tan dog with a white wolf of similar size. Preserve the dog's original run, jump, landing and timing. Keep the crouching person and snowy surroundings unchanged.
\begin{figure}[H]\centering
\includegraphics[page=16,width=.97\linewidth]{supplement_figures/panels.pdf}
\caption[Supplementary figure 16]{\input{supplement_figures/caption_s16.tex}}\label{fig:s16}
\end{figure}

\clearpage
\noindent\textbf{Small targets and shaded surfaces.}\par
\noindent\textbf{Case 16: Small car in a moving aerial view.} Change only the light-colored car closest to the center of the intersection to bright red. Preserve the car's shape and trajectory, all other vehicles, road markings and buildings.
\begin{figure}[H]\centering
\includegraphics[page=17,width=.97\linewidth]{supplement_figures/panels.pdf}
\caption[Supplementary figure 17]{\input{supplement_figures/caption_s17.tex}}\label{fig:s17}
\end{figure}
\noindent\textbf{Case 17: Red leather in seat shadows.} Change only the dark upholstery of the driver's seat to red leather. Preserve the driver, their clothing, the steering wheel, window views and vehicle motion.
\begin{figure}[H]\centering
\includegraphics[page=18,width=.97\linewidth]{supplement_figures/panels.pdf}
\caption[Supplementary figure 18]{\input{supplement_figures/caption_s18.tex}}\label{fig:s18}
\end{figure}

\clearpage
\subsection{Additional editing tasks}
\begin{figure}[H]\centering
\includegraphics[page=19,width=.97\linewidth]{supplement_figures/panels.pdf}
\caption[Supplementary figure 19]{\input{supplement_figures/caption_s19.tex}}\label{fig:s19}
\end{figure}
\clearpage
\noindent\textbf{Object and background replacement.}
\begin{figure}[H]\centering
\includegraphics[page=20,width=.97\linewidth]{supplement_figures/panels.pdf}
\caption[Supplementary figure 20]{\input{supplement_figures/caption_s20.tex}}\label{fig:s20}
\end{figure}
\clearpage
\noindent\textbf{Appearance, material and style.}
\begin{figure}[H]\centering
\includegraphics[page=21,width=.97\linewidth]{supplement_figures/panels.pdf}
\caption[Supplementary figure 21]{\input{supplement_figures/caption_s21.tex}}\label{fig:s21}
\end{figure}
\clearpage
\noindent\textbf{Object addition.}
\begin{figure}[H]\centering
\includegraphics[page=22,width=.97\linewidth]{supplement_figures/panels.pdf}
\caption[Supplementary figure 22]{\input{supplement_figures/caption_s22.tex}}\label{fig:s22}
\end{figure}
\subsection{Shared failure on a moving person}
\noindent\href{https://viva-paper.github.io/assets/comparison/ins/073/src.mp4}{VIVA 073.} Remove the boy.
\begin{figure}[H]\centering
\includegraphics[page=23,width=.97\linewidth]{supplement_figures/panels.pdf}
\caption[Supplementary figure 23]{\input{supplement_figures/caption_s23.tex}}\label{fig:s23}
\end{figure}
\noindent\textbf{Failure analysis.} Changing scale, pose and ball interaction complicate target tracking and background reconstruction; these are plausible challenges, not verified causes. TriDiT accelerates the backbone without explicitly correcting instruction-following failures. \textbf{Scope of the advantage.} The benchmark runtime gain does not remedy this failure; no quality advantage is established here.

%% file: supplement_figures/caption_s09.tex
Timing (seconds): DiT 126.48 to 50.80 (2.49$\times$); attention core 75.89 to 22.01 (3.45$\times$). E2E speedup: 2.25$\times$. E2E adds 9.60 s of non-DiT overhead.

%% file: supplement_figures/caption_s10.tex
Timing (seconds): DiT 118.76 to 46.57 (2.55$\times$); attention core 71.26 to 19.95 (3.57$\times$). E2E speedup: 2.31$\times$. E2E adds 8.40 s of non-DiT overhead.

%% file: supplement_figures/caption_s11.tex
Timing (seconds): DiT 94.35 to 51.00 (1.85$\times$); attention core 56.61 to 22.64 (2.50$\times$). E2E speedup: 1.74$\times$. E2E adds 7.80 s of non-DiT overhead.

%% file: supplement_figures/caption_s12.tex
Completion. The selected woman is absent from all seven TriDiT samples. Temporal consistency. The vacated left-hand region remains free of the target across the displayed times. Preservation. The foreground curly-haired woman remains, but other walkers and their arrangement change; successful removal therefore coexists with collateral changes. Columns align by playback time. TriDiT advantage. TriDiT completes selective removal while retaining the main foreground figure; the demonstrated strength is target completion, with background preservation still imperfect.

%% file: supplement_figures/caption_s13.tex
Completion. The bowl remains in the sampled Lucy-Edit-Dev, Ditto and InsV2V outputs, while the boxed countertop region is empty in all six TriDiT samples. Temporal consistency. No reappearance of the bowl is visible at these times. Preservation. TriDiT broadly retains the kitchen layout and foreground pastry. Ditto also changes the kitchen framing and appearance while retaining the bowl. These observations distinguish completing the removal from simply producing a plausible kitchen. Latency is summarized separately in Table S6. TriDiT advantage. Unlike the displayed baselines, TriDiT removes the small object without losing the main kitchen layout, combining completion with coarse preservation.

%% file: supplement_figures/caption_s14.tex
Completion. TriDiT places a white dog beside the woman. Ditto and InsV2V leave the sofa empty; Lucy-Edit-Dev produces an oversized, horse-like white animal. Temporal consistency. The dog remains present across the six TriDiT samples while its pose changes. Preservation. The woman's position, dark wall, sofa and foreground table broadly follow the source layout. This supports coarse scene preservation, but does not imply identical facial details or motion between sampled times. Boxes identify the added-object region. TriDiT advantage. TriDiT better matches the requested object category and placement than the displayed baselines, with the added dog persisting across samples.

%% file: supplement_figures/caption_s15.tex
Completion. Both methods turn the selected jacket red across the seven samples. Temporal consistency. The red appearance persists as the sleeve and hands move. Preservation. The enlarged cuff retains a visible boundary between the red sleeve, bare hand and neighboring light clothing. The other people's clothing remains broadly consistent with the source. These frames support localized recoloring, without establishing pixel-exact boundary preservation. TriDiT advantage. Compression retains the visible localized recoloring and cuff separation of the uncompressed reference, illustrating preservation of edit selectivity.

%% file: supplement_figures/caption_s16.tex
Completion. Once the animal enters view, both outputs show a pale, wolf-like replacement; the small frames alone cannot establish fine-grained species fidelity. Temporal consistency. The displayed sequence retains the broad approach and passage beside the crouching person. Exact footfall, jump and landing timing require denser frames. Preservation. Snow, trees and the person's broad placement remain recognizable, while animal proportions and pose details differ from the source. TriDiT advantage. TriDiT preserves a persistent replacement appearance and the broad action sequence despite source-token compression.

%% file: supplement_figures/caption_s17.tex
Completion. The enlarged target is red in both outputs. Temporal consistency. The moving aerial sequence retains a colored vehicle, although its small size limits reliable identity tracking in every sample. Preservation. The MiniMax-H3-ref2VA crop changes the intersection orientation. TriDiT more closely retains the source crosswalk orientation and adjacent white vehicle in the shown crop. This is localized evidence of background preservation, rather than proof of superiority throughout the video. TriDiT advantage. The shown crop combines small-target recoloring with closer preservation of nearby road geometry and the non-target vehicle.

%% file: supplement_figures/caption_s18.tex
Completion. Both outputs introduce red upholstery. At the enlarged time, MiniMax-H3-ref2VA concentrates red near the rim; TriDiT extends a darker burgundy tint across the shaded bolster. Temporal consistency. Red seat regions persist across the seven samples. Preservation. The rear panel remains darker and the driver and steering-wheel layout remain broadly recognizable. The tint is visually compatible with shading, but the frames do not verify physical lighting accuracy. TriDiT advantage. TriDiT recolors more of the visible shaded bolster while retaining tonal separation, rather than limiting red to the bright rim.

%% file: supplement_figures/caption_s19.tex
01. Both outputs retain the drone and broad shoreline layout across samples; the small shadow is unresolved. 02. Luminous smoke develops above the spout, but both outputs also darken the surroundings. 03. The boxed regions show water rather than a visible hull, although target size limits certainty. Sea texture and horizon appearance also change. 04. The boat is absent and riverbanks remain broadly aligned. Fine wake remnants and water continuity need denser inspection. TriDiT advantage. Cases 01 and 04 show persistent addition and removal with recognizable scene layout under compression; they do not establish superiority for every task.

%% file: supplement_figures/caption_s20.tex
05. A sailing ship replaces the cargo vessel; neighboring boats remain visible. 06. Red machines persist, but the person's pose also changes. 07. Green vegetation replaces the landscape; foreground subjects remain visible, with fine limb continuity unverified. 08. Snowy trees replace the waterfall; the woman's silhouette remains. TriDiT advantage. Cases 05 and 07 combine replacement with retention of nearby vessels or foreground subjects under compression.

%% file: supplement_figures/caption_s21.tex
09. Blue wax persists across seven samples in both outputs, while the warm flame and dark background remain distinct. 10. Both mugs show the desk tone through the upper wall at the three enlarged times, supporting apparent transparency. The reference has repeated lower-wall highlights; TriDiT has a broader pale base patch. Glass-like appearance persists, without establishing physically accurate refraction or exact geometry. The man, monitor and papers retain their broad layout. 11. The oil-painting result shows brushwork across the scene while retaining recognizable rooflines and windows. TriDiT advantage. Cases 09-10 retain persistent color and material changes under compression, including a transparent-looking mug with recognizable surrounding objects.

%% file: supplement_figures/caption_s22.tex
12. A bird is added above the wires. 15. The red balloon and broad coastal layout persist across samples; precise world-space anchoring and between-frame motion remain unverified. TriDiT advantage. Case 15 retains the added balloon and coastal layout across the displayed times under compression.

%% file: supplement_figures/caption_s23.tex
Completion fails: Lucy-Edit-Dev and InsV2V retain the boy in all seven samples. Continued visibility does not indicate successful removal. The same failure is author-reported for MiniMax-H3-ref2VA and TriDiT.